\newif\ifdraft
\draftfalse

\documentclass[preprint,12pt]{elsarticle}

\usepackage{amssymb}
\usepackage{amsmath}

\usepackage{algorithm,algorithmic}
\usepackage{amsfonts}
\usepackage{hyperref}
\usepackage{bbm}
\usepackage{graphicx}
\usepackage{stfloats}  
\usepackage{caption,subcaption}
\usepackage{multirow}
\usepackage{cleveref}
\usepackage{comment}
\usepackage{hhline}
\usepackage{todonotes}
\usepackage{siunitx}
\usepackage{booktabs}
\usepackage{csvsimple}
\usepackage{datatool}
\usepackage{array}
\usepackage{xcolor}
\usepackage[normalem]{ulem}
\hypersetup{hidelinks=true}
\usepackage{textcomp}

\newcommand{\header}[1]{\noindent\textbf{#1:}}

\usepackage[normalem]{ulem} 
\newcommand{\stkout}[1]{\ifmmode\text{\sout{\ensuremath{#1}}}\else\sout{#1}\fi}

\ifdraft

\newcommand{\deleted}[1]{\textcolor{red}{\stkout{#1}}}

\newcommand{\deletedfloat}[1]{}
\newcommand{\commented}[1]{\textcolor{blue}{#1}}
\else

\newcommand{\deleted}[1]{}

\newcommand{\deletedfloat}[1]{}
\newcommand{\commented}[1]{}
\fi

\journal{Computers in Biology and Medicine}

\begin{document}

\begin{frontmatter}



\title{S4Sleep: Elucidating the design space of deep-learning-based sleep stage classification models}


\author{Tiezhi Wang} 

\affiliation{organization={Carl von Ossietzky Universität Oldenburg},
            addressline={Ammerlaender Heerstr. 114-118}, 
            city={Oldenburg},
            postcode={26129}, 
            state={Lower Saxony},
            country={Germany}}
            
\author{Nils Strodthoff\corref{cor1}} 

\affiliation{organization={Carl von Ossietzky Universität Oldenburg},
            addressline={Ammerlaender Heerstr. 114-118}, 
            city={Oldenburg},
            postcode={26129}, 
            state={Lower Saxony},
            country={Germany}}
\ead{nils.strodthoff@uol.de}

\begin{abstract}

Machine-learning-based automatic sleep stage scoring is a promising approach to enhance the time-consuming manual annotation process of polysomnography recordings. Although numerous algorithms have been proposed for this purpose, systematic exploration of architectural design decisions remains limited. This study conducts a comprehensive investigation into these design choices within the broad category of encoder-predictor architectures. The methodology identifies robust architectures applicable to both time series and spectrogram input representations, both of which leverage structured state space models as integral components. Without further hyperparameter adjustments, the proposed models S4Sleep(spec) and S4Sleep(ts) consistently surpass all existing approaches on the most commonly used benchmark datasets: Sleep EDF, the Montreal Archive of Sleep Studies, and, most notably, the extensive Sleep Heart Health Study dataset. The architectural insights derived from this research, along with the refined methodology for architecture search demonstrated herein, are expected to not only advance future research in sleep staging but also be beneficial for other time series annotation tasks.

\end{abstract}


\end{frontmatter}



\section{Introduction}
\label{sec:introduction}

Sleep disorders, which are prevalent and can have severe implications for patients' health and overall well-being, are known to lead to significant morbidity \cite{panossian2009review, pavlova2019sleep}. Primary care practice commonly involves managing various major sleep disorders, including insomnia, sleep-disordered breathing, hypersomnia/narcolepsy, circadian rhythm disorders, parasomnias, and sleep-related movement disorders \cite{pavlova2019sleep}. Within clinical practice, sleep staging is crucial for assessing the structure and quality of sleep, enabling the diagnosis of sleep disorders and informing appropriate treatment strategies. It also facilitates sleep research, providing insights into sleep patterns and physiological processes during different stages of sleep.

\header{Sleep staging}
Sleep staging refers to categorizing the state of human sleep into one of five or six sleep stages: wake stage (W), non-rapid-eye-movement (NREM) stages (N1-N3 or N1-N4 depending on the annotation standard), and rapid-eye-movement (REM) stage. The gold standard measurement setup for sleep staging is polysomnography (PSG), which typically comprises a range of different sensor modalities, such as electroencephalography (EEG), electrooculogram (EOG), electromyography (EMG), electrocardiography (ECG), pulse oximetry, and respiratory signals. The EEG is the most critical modality; therefore, most single-channel automatic sleep staging algorithms rely on a single representative EEG channel. The configurations of EEG montages and the availability of other sensor modalities vary widely across datasets. The annotation process is formalized through the American Academy of Sleep Medicine (AASM) guidelines \cite{iber2007aasm} or the Rechtschaffen and Kales (R\&K) guidelines \cite{hobson1969manual} and is usually performed for 30-second segments, referred to as epochs. The assessment of an overnight PSG recording by a human expert is a tedious process, requiring up to several hours of work, and is subject to considerable inter-rater variability \cite{DANKERHOPFE2009,Rosenberg2013}.

\header{Prospects of automatic sleep staging}
A study \cite{stephansen2018neural} evaluated machine-learning-based automated sleep staging models on PSG recordings scored by multiple human raters, with the consensus of these raters serving as the benchmark. The results demonstrated that automatic sleep staging not only significantly outperformed human raters in terms of accuracy but also exhibited higher Cohen's kappa values, indicating superior inter-rater reliability, i.e., a better alignment with the consensus of human raters. Even more striking, a recent study \cite{fiorillo2023u} showed that a deep learning-based algorithm for sleep staging, even when trained on datasets characterized by noisy labels produced by various human raters adhering to the AASM guidelines, can achieve similar performance even when trained on non-standard channel configurations and thus does not require strict compliance with the AASM standards to achieve reliable performance. In fact, the inclusion of automatic sleep staging methods as part of the assessment protocol could mark a long-overdue step towards enhancing the standardization of the assessment procedure to reduce both inter-rater and inter-center variability \cite{fiorillo2023u}. This provides a clear incentive for the identification of robust and increasingly accurate model architectures that generalize across datasets and input configurations.

\header{Rationale for this study}
While the field of automatic sleep staging using machine learning has evolved significantly in recent years, compelling evidence on several important design decisions is lacking. First, there is no consensus on the most suitable input representation. State-of-the-art models from the literature typically rely on spectrograms as the input representation, which, however, unlike raw time series, do not retain the full complexity of the input signal. Even though spectrograms encode a useful inductive bias, competitive models trained on raw time series have the potential to outperform spectrogram-based models when (pre-)trained on large datasets due to their ability to learn more nuanced representations. Second, even for a given input representation, architectural design choices have rarely been investigated systematically. Most notably, beyond the predominantly used recurrent neural networks \cite{hochreiter1997long} and, more recently, transformer models \cite{vaswani2017attention}, such an investigation should include structured state space sequence (S4) models, which are well known for their ability to capture long-range dependencies in time series data \cite{Gu2021EfficientlyML} and have been successfully applied to other physiological time series data, such as electrocardiography data \cite{Mehari2023S4}. This work addresses the mentioned gaps in the research landscape and proposes model architectures that achieve new state-of-the-art performance for automatic sleep staging on various commonly used datasets.

\header{Technical contributions}
The main contributions of this work can be summarized as follows:\\
(1) Using sleep staging as a case study, this research devises a systematic procedure to identify optimal model architectures for long time series annotation tasks within the search space of encoder-predictor architectures, which encompasses most literature approaches.\\
(2) Building on this procedure, this study identifies optimal model architectures for both raw time series and spectrograms as input, as well as for single-channel and multi-channel configurations, which outperform literature approaches on the Sleep EDF dataset and the Montreal Archive of Sleep Studies dataset—two widely used public medium-sized benchmark datasets for sleep staging—and, most notably, on the large-scale Sleep Heart Health Study dataset when trained without any further hyperparameter adjustments. This underscores the robustness of the study's findings.\\
(2.1) For both raw time series and spectrogram input representations, the study identifies structured state space models as highly effective sequence encoders, albeit with input-modality-specific preprocessing requirements.\\
(2.2) Optimal predictor models aggregating epoch-level predictions were also found to be structured state space models, irrespective of the input representation under consideration.

\header{Organization of this paper}
The remainder of the manuscript is organized as follows: Sec.~\ref{sec:relatedwork} reviews existing approaches under the unifying umbrella of encoder-predictor models, which constitutes the broad class of model architectures under investigation. Sec.~\ref{sec:methods} introduces the modular approach to model construction, starting from single-epoch prediction and extending to multi-epoch prediction models. Furthermore, it introduces the considered datasets, training methodology, as well as target metrics and evaluation procedures. Sec.~\ref{sec:results} first presents the results of the process that led to the identification of the S4Sleep(ts) and S4Sleep(spec) model architectures and, in a second step, evaluates these models on a large set of benchmark datasets to demonstrate the robustness of the study's findings. In Sec.~\ref{sec:discussion}, these results are contextualized in a comprehensive discussion, addressing limitations and future research perspectives. Finally, the findings are summarized in Sec.~\ref{sec:summary}. Additional results and technical details are presented in the supplementary material.

\section{Related Work}
\label{sec:relatedwork}

Machine learning, particularly deep learning algorithms, has improved tremendously in the past few years. Recently, sleep staging algorithms have achieved a level of performance comparable to human inter-rater variability \cite{stephansen2018neural}. While most approaches still follow a purely supervised training procedure, self-supervised pretraining is increasingly considered to learn representations from unlabeled data \cite{zhang2022self,gorade2023large} before finetuning on labeled data. Models operate on time-domain, frequency-domain, or hybrid input representations \cite{phan2021xsleepnet}. The sleep staging task is typically framed as a sequence task, where multiple epochs—i.e., 30-second segments, of input signal in the chosen input representation are mapped to a sequence of sleep stage annotations, one per input epoch.

\begin{figure}[h]
	\centering
		\includegraphics[width=0.99\linewidth]{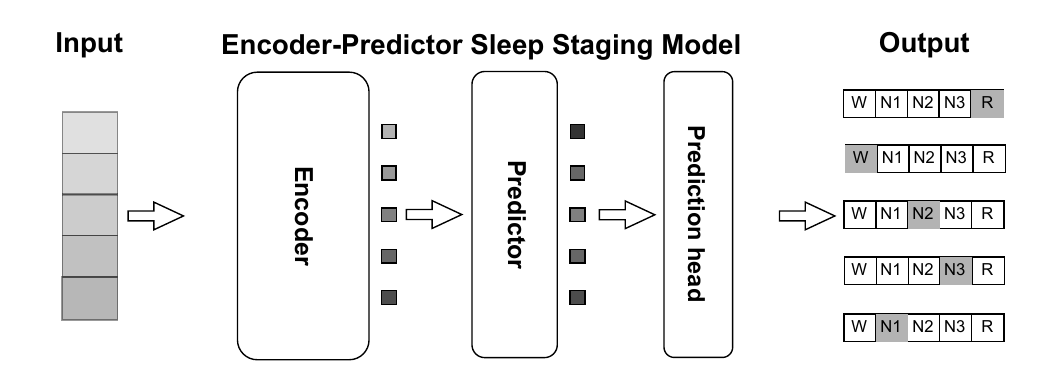}
	\caption{Schematic representation of the encoder-predictor architecture used in sleep staging models. }
	\label{fig:arch}
\end{figure}

\header{Encoder-predictor models} 
Most literature approaches can be categorized as encoder-predictor models. The encoder extracts local features, which are typically reduced to one output token per input epoch, while the predictor aggregates local features according to the temporal context \cite{phan2019seqsleepnet}, see \Cref{fig:arch} for a schematic representation. Such architectures are also preferable on general grounds as they adhere to the principle of scale separation \cite{bronstein2021geometric} and facilitate hierarchical representations \cite{schmidhuber1992learning,lecun2022path}.

\header{Unifying literature approaches} 
In \Cref{tab:architectures}, literature approaches are categorized according to the model components utilized. First, the table demonstrates that the class of encoder-predictor models is a broad and inclusive category encompassing the majority of literature approaches. The most notable model architecture from the literature that does not fall into the category of encoder-predictor models in the above sense is the U-Sleep model \cite{Perslev2021}, which instead relies on a U-Net architecture, as well as single-epoch models such as \cite{an2022amplitude,pei2024automatic}. Second, encoder and predictor models are often chosen identically, as seen in models such as (L-)SeqSleepNet \cite{phan2023lseqsleepnet}, SleepTransformer \cite{phan2022sleeptransformer}, or RobustSleepNet \cite{guillot2021robustsleepnet}. The predominant choice for these models has historically been recurrent neural networks \cite{phan2019seqsleepnet,phan2021xsleepnet,guillot2021robustsleepnet,phan2023lseqsleepnet}, while more recently, transformer models have also been explored \cite{phan2022sleeptransformer}. The pooling operation to reduce the temporal resolution of the encoder representation is often realized through an attention mechanism.

\begin{table}[h]
	\centering
	\caption{Unified encoder-predictor approaches for sleep staging, incorporating both literature methods and the models proposed in this work.}
	\label{tab:architectures}	
	\resizebox{\textwidth}{!}{\begin{tabular}{llllll}
		\toprule
		model & input & encoder & pooling & predictor & remarks\\\midrule
		\textit{S4Sleep(ts)} & ts & CNN+S4 & average & S4 & this work\\
		\textit{S4Sleep(spec)} & spec & S4 & average & S4 & this work\\ 	
		L-SeqSleepNet\cite{phan2023lseqsleepnet}&spec&LSTM&folding&LSTM&long sequence input\\
		ProductGraphSleepNet\cite{einizade2023productgraphsleepnet} &de&\multicolumn{2}{l}{GCN (as both encoder and pooling)}&GRU&graph-based\\
		SAGSleepNet \cite{jin2023sagsleepnet} &spec&GCN&average&GRU&graph-based\\
		SleepTransformer\cite{phan2022sleeptransformer}&spec&Transformer&attention&Transformer&\\
		XSleepNet1 \cite{phan2021xsleepnet}&ts+spec&ts: CNN spec: FB&attention&ts: GRU spec: LSTM&dual-modality input\\
		XSleepNet2 \cite{phan2021xsleepnet}&ts+spec&ts: CNN spec: FB&attention&ts: GRU spec: LSTM&dual-modality input\\
		RobustSleepNet \cite{guillot2021robustsleepnet}&spec&channel-attention + GRU&attention&GRU&\\		
		IITNet \cite{seo2020intra}&ts&ResNet&-&LSTM&sub-epoch encoder\\
		SeqSleepNet \cite{phan2019seqsleepnet}&spec&FB+LSTM&attention&LSTM&\\
		DeepSleepNet \cite{supratak2017deepsleepnet}&ts&CNN&max&LSTM&\\
		\bottomrule	
	\end{tabular}}
	\captionsetup{justification=centering, singlelinecheck=false, position=top}
	\parbox{\linewidth}{\tiny ts: raw time series. spec: spectrograms. de: differential entropy. GCN: graph convolutional network. FB: filter bank layer composed of a restricted fully connected layer. folding: folding subnetwork}
\end{table}

\section{Methods}
\label{sec:methods}

\subsection{Models}
\header{Encoder-Predictor Model} 
This study follows the design choice of encoder-predictor models introduced above, which is described here at a higher level of detail: Roughly speaking, the  input (either raw time series or spectrograms) is processed by an encoder model, which typically returns a semantically enriched, temporally downsampled representation. The conventional approach in the field is to reduce the temporal resolution to a single token per sleep staging epoch. However, it is also possible to choose a temporally more fine-grained representation at the sub-epoch level \cite{seo2020intra} for the intermediate representation. 
Consequently, this representation is passed to the predictor model, which integrates the information across tokens on the epoch or sub-epoch level. Finally, the output of the predictor model is processed by a prediction head to produce epoch-level predictions. In the case of an epoch-level encoder, prediction head can be as simple as a MLP or a single linear layer shared across all output tokens. In the case of a sub-epoch-level encoder, a local pooling layer is typically applied beforehand to reach an output representation that matches the required temporal resolution at the epoch level.

\header{Modular design} 
Different encoder and predictor architectures can be combined in a modular fashion. One of the core contributions of this work is to provide a structured assessment of their different components and their interplay. Beginning with an exploration of single-epoch prediction models facilitates the identification of superior architectural designs adept at abstracting high-quality, epoch-level representations. This initial step is crucial for pinpointing encoder architectures that are highly effective for multi-epoch models. Subsequently, long-range dependencies spanning across multiple epochs can be captured through a prediction model operating on (sub-)epoch-level compressed representations. Various specific instantiations of the prediction model can be considered. The following paragraphs provide concise descriptions of the different architectural components considered in the experiments to maintain the flow of reading. More detailed descriptions are provided in ~\ref{app:architectures} in the supplementary material and in the accompanying code repository \cite{repo}.

\subsubsection{Single-epoch prediction models}

\header{Single-epoch encoders}
As single-epoch encoder, this research considers shallow CNN encoders or directly passing the input to the predictor. Using a shallow CNN is a commonly adopted design choice for both raw waveforms and spectrogram-based audio processing, see for example \cite{baevski2020wav2vec,radford2023robust}, as it allows mapping the raw input to a more favorable input representation for the predictor while adding little complexity and optionally (in cases of strided convolutions) reducing the temporal resolution of the signal. The CNN encoder for spectrogram input can be seen as a generalization of the Whisper encoder \cite{radford2023robust} to multiple input channels. The following section lists the considered encoder modules along with their abbreviations for later reference. Refer to ~\ref{app:sing-ep} for a more detailed description.

\begin{itemize}
	\item time series input:
	\begin{itemize}
		\item \textit{CNN}: two layers of strided 1D convolutions interleaved with ReLU activations
		\item \textit{NONE}: input directly passed to the predictor
	\end{itemize}
	\item spectrogram input:
	\begin{itemize}
		\item \textit{CNN}: reshaping of frequency and channel axes into a single axis, followed by two layers of 1D convolutions interleaved with GeLU activations
		\item \textit{NONE}: reshaping of frequency and channel axes into a single axis
	\end{itemize}
\end{itemize}

\header{Single-epoch predictors} 
This study considers a bidirectional LSTM model as a representative for the predominantly used recurrent neural network predictors, in addition to transformer and structured state space sequence (S4) models. Again, the reader is referred to ~\ref{app:sing-ep} for a more detailed description.

\begin{itemize}
	\item \textit{LSTM}: two-layer bidirectional LSTM model
	\item \textit{TF}: four-layer transformer model 
	\item \textit{S4}: four-layer S4 model
\end{itemize}

\header{Single-epoch prediction heads} 
To keep the scope of the study limited, this study only explores global/local average pooling layers followed by a linear layer with five output neurons.

\subsubsection{Multi-epoch prediction models}

\header{\\Multi-epoch encoders} 
For multi-epoch input, this study considers different strategies. First, it examines the so-called epoch encoder, corresponding to identified optimal single-epoch prediction models (\textit{EES4} and \textit{EENS4}). 

In the case of raw time series input, the study also considers a shallow CNN with large strides to provide a temporally downsampled hidden representation (\textit{SCNN}). Spectrograms as input modality are already heavily downsampled compared to the original time series input. This allows the direct use of the encoders from the single-epoch case (\textit{CNN} and \textit{NONE}) without further internal aggregation. As before, the reader is referred to \ref{app:multi-ep} for technical details.

\begin{itemize}
	\item time series input:
	\begin{itemize}
		\item \textit{EES4}: \textit{CNN} encoder with \textit{S4} predictor and average pooling as an epoch encoder
		\item \textit{SCNN}: four layers of strided 1D convolutions interleaved with ReLU activations with a temporal downsampling factor of 100, inspired by the Wav2vec 2.0 encoder for raw audio data \cite{baevski2020wav2vec}
	\end{itemize}
	\item spectrogram input:
	\begin{itemize}
		\item \textit{EENS4}: \textit{NONE} encoder with \textit{S4} predictor and average pooling as an epoch encoder
		\item \textit{CNN}: using the single-epoch CNN encoder for spectrograms as input without any aggregation on the epoch level
		\item \textit{NONE}: using \textit{NONE}, defined as the single-epoch encoder		
	\end{itemize}
\end{itemize}

\header{Multi-epoch predictors}
This study considers the same predictors as for the single-epoch prediction models.

\header{Multi-epoch prediction heads} 
This study considers the same prediction heads as for the single-epoch prediction models.

\subsection{Datasets}

\header{Considered datasets} 
This study bases its analysis on three widely used datasets for automatic sleep staging. The Sleep EDF (SEDF) dataset \cite{kemp2000analysis, goldberger2000physiobank} forms the basis for the model selection experiments. The study evaluates the performance on a hold-out set unseen in this model selection procedure. Due to its widespread use in the literature, the authors also consider the smaller SEDF20 subset of SEDF. In addition to SEDF, The authors consider a particular subset of the Montreal Archive of Sleep Studies (\textit{MASS-SS3}) \cite{OReilly2014} and the first visit within the large-scale Sleep Heart Health Study (\textit{SHHS1}) datasets \cite{ross2019probabilistic, quan1997sleep}. The authors summarize basic characteristics of the considered datasets in \Cref{tab:datasets}. In \ref{app:datasets} of the supplementary material, a detailed description of the dataset preparation procedures and the rationale for selecting these datasets are provided.

\begin{table}[h]
	\centering
	 \scriptsize
	\caption{Summary of datasets used in this study.}
	\label{tab:datasets}
	\begin{tabular}{p{3.5cm}p{2cm}p{1.5cm}p{4.9cm}}
	\hline
	dataset & recordings & patients & included channels \\ \hline
	SEDF(full) \cite{kemp2000analysis, goldberger2000physiobank} & 197 & 100 & EEG (Fpz-Cz${}^\ast$, Pz-Oz), EOG (horizontal) \\
	SEDF20 \cite{kemp2000analysis, goldberger2000physiobank} & 39& 20& EEG (Fpz-Cz${}^\ast$, Pz-Oz), EOG (horizontal) \\
	MASS-SS3 \cite{OReilly2014}& 62 & 62 & EEG (20 channels${}^\dag$) EOG (left, right), EMG, ECG \\
	
	SHHS Visit 1 \cite{ross2019probabilistic, quan1997sleep} & 5463&5463 & EEG (C3-A2, C4-A1${}^\ast$), EOG (left, right), EMG \\ \hline
	\end{tabular}
	\parbox{\linewidth}{\tiny ${}^\ast$: selected channel for single-channel configuration experiments. ${}^\dag$: F4-EOG obtained via montage reformatting \cite{lagerlund2000manipulating} for single-channel configuration experiments}
\end{table}

\header{Preprocessing} All datasets were resampled to 100~Hz using bandlimited interpolation \cite{Smith2015DigitalAudioResampling}. In all cases, this study considered two input representations: raw time series, where the signal underwent no further preprocessing, and spectrograms, which were generated in accordance with with \cite{phan2023lseqsleepnet}. Details are provided in \ref{app:datasets} of the supplementary material. 

\subsection{Training procedure and performance evaluation}
\label{sec:train_and_eval}
\header{Training procedure}
This study used the focal loss \cite{lin2017focal} as the loss function to mitigate issues arising from imbalanced label distributions. Optimization was performed using AdamW optimizer \cite{loshchilov2018decoupled} at a constant learning rate. Model selection was performed on the validation set to select the best model for later evaluation. During test time, this study averaged predictions from sequentially moving the input through the entire sequence in a sliding window fashion. The reader is referred to \ref{app:train_and_eval} in the supplementary material for a more detailed description of the training procedure.

\header{Performance metrics, score estimation and predictive uncertainty} By convention in the field of sleep staging, this study focuses on the macro-$F_1$ score, i.e., the mean of the individual label $F_1$-scores, as the primary performance metric. This metric is complemented by the macro average of the area under the receiver operating curve (AUROC) as a ranking metric operating on continuous outputs as opposed to dichotomized predictions in the case of the macro-$F_1$ score. This study used a holdout set evaluation scheme for the larger SEDF(full) and SHHS1 datasets and a cross-validation scheme for SEDF20 and MASS-SS3, as conventionally done in the literature. The study assessed systematic uncertainties arising from the randomness of the training process by conducting multiple training runs. Additionally, statistical uncertainties and significance were evaluated through bootstrapping on the test set. For additional details, the reader is referred to \ref{app:train_and_eval} of the supplementary material.

\section{Results}
\label{sec:results}

\begin{figure*}[!!!h]
	\centering
	\begin{subfigure}[b]{0.49\textwidth}
		\centering
		\includegraphics[width=\linewidth]{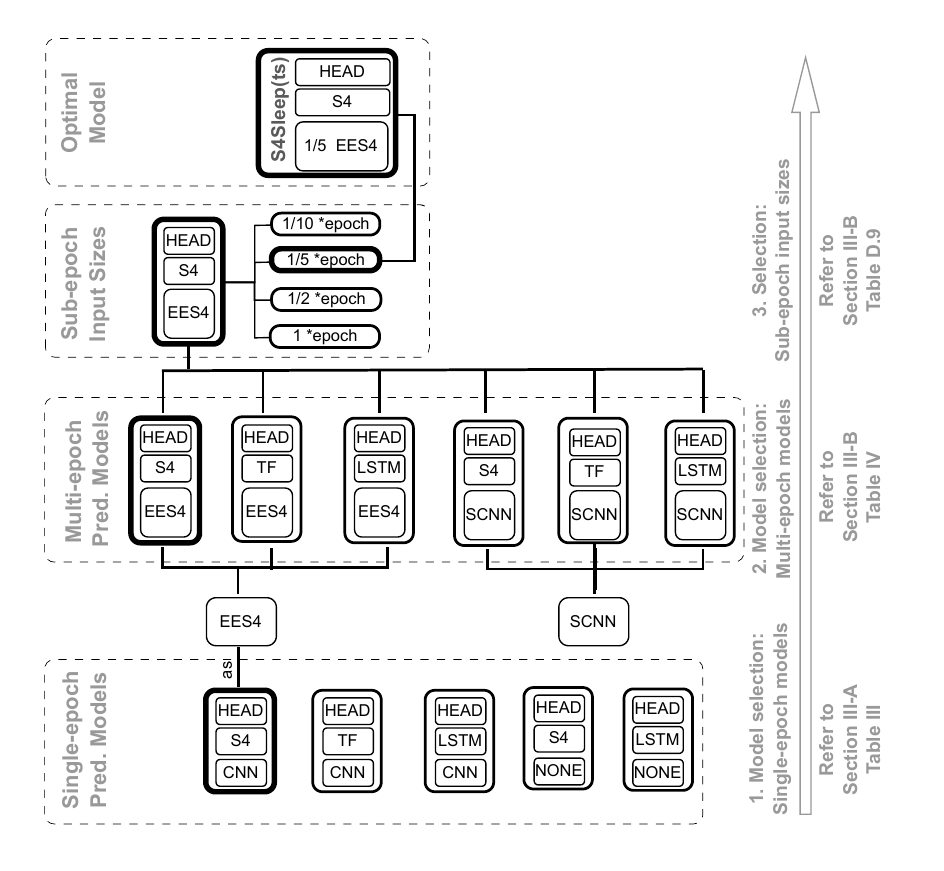}
		\caption{Raw time series as model input}
		\label{fig:chart_ts}
	\end{subfigure}
	  \hfill
	\begin{subfigure}[b]{0.49\textwidth}
		\centering
		\includegraphics[width=\linewidth]{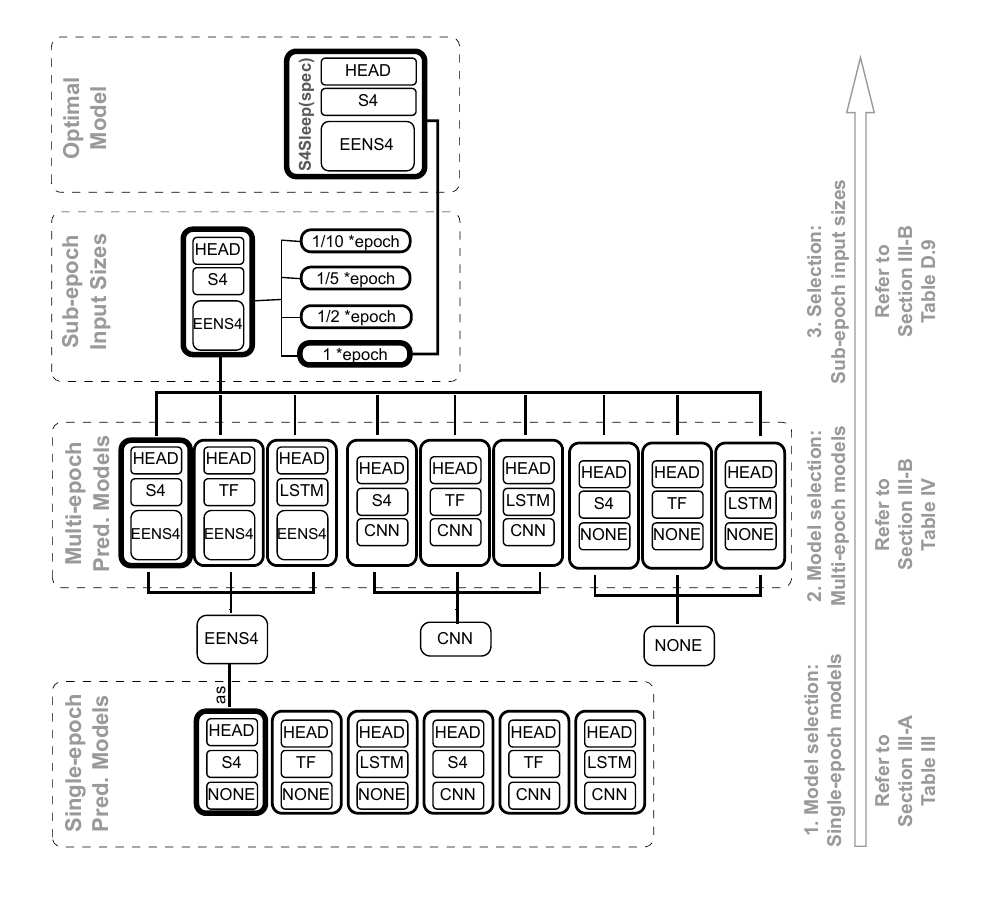}
		\caption{Spectrogram as model input}
		\label{fig:arch2}
	\end{subfigure}
	
	\caption{Flow chart demonstrating the organization of the experiments that led to the identification of optimal model architectures for (a) raw time series and (b) spectrograms as input. Both start by identifying strong single-epoch models, which are then leveraged (in the form of epoch encoders) to explore multi-epoch prediction models while also investigating alternative encoder choices. Subsequently, for the two identified architectures the usage of sub-epoch instead of full-epoch encoders is explored. Finally, the two best-performing model architectures, \textit{S4Sleep(ts)} and \textit{S4Sleep(spec)}, are evaluated on held-out test sets and retrained on the commonly used SEDF20 datatset , the MASS-SS3 dataset and the large-scale SHHS1 dataset, culminating in the final performance evaluation compiled in \Cref{tab:mainresult_sedf} and \Cref{tab:mainresult_massshhs}.}
	\label{fig:chart}
 \end{figure*}

\header{Organization of experiments} The course of the experiments is summarized in Figure~\ref{fig:chart}.
The study proceeded in the following stages: To restrict the search space, the study started with a first stage as described in \Cref{sec:results_single} by considering models that take a single epoch as input. Based on the insights gained from this experiment, the research then proceeded as described in \Cref{sec:results_multi} to evaluate multi-epoch models leveraging the optimal model architectures identified in the first stage as epoch encoders. In this second stage, the goal was to identify optimal encoder and predictor combinations for both  raw time series and spectrograms as input. Finally, the study investigated whether the identified models can benefit from sub-epoch as opposed to full-epoch encoders. To prevent overfitting to the SEDF(full) test set, only validation set performances were reported at this stage. The study concludes the results part with \Cref{sec:testseteval}, where it puts the identified model architectures to the test on two different subsets of SEDF. To validate the robustness of the findings, this study also tested the identified model architectures on two additional datasets: the middle-scale MASS-SS3 dataset and the large-scale SHHS1 dataset.

\subsection{Model selection: single-epoch models}
\label{sec:results_single}
For single-epoch models, this study assessed the impact of the predictor models (\textit{S4}, \textit{TF}, \textit{LSTM}). For both input representations, the authors considered convolutional encoders (\textit{CNN}) as well as no encoder (\textit{NONE}). All results are compiled in \Cref{tab:sedf1}. The results for raw time series as input for both single- and multi-channel setups clearly distinguish \textit{CNN+S4} as the  best-performing architecture. For time series input, a CNN encoder proved to be beneficial as compared to using no encoder. Conversely, for spectrogram data, using no encoder emerges as the more favorable choice. These results suggest that the architectural choices are inherently tied to the input representations. The difference between the two encoder choices for an identical predictor model is most strongly pronounced for S4 as a predictor, whereas TF is minimally influenced by this choice. The best-performing model on spectrogram input according to the macro-$F_1$ score is \textit{NONE+S4}, again across both channel settings. Looking at the overall best performance, raw-time-series-based models are outperformed by spectrogram-based models for multi-channel input while the opposite ordering is observed in the single-channel case. In line with expectations, multi-channel input consistently exhibits a systematic positive effect compared to single-channel input. This study used these insights from the single-epoch case to construct multi-epoch prediction models using the identified encoder-predictor combinations as epoch encoders.

\begin{table}[!!!h]
	\centering
	\scriptsize
	\caption{Comparison of different encoder and predictor architectures for a \textit{single epoch input size} trained on SEDF based on validation set scores. Best results for a given channel configuration are marked in boldface and demonstrate statistically significant superiority over all competing alternatives. The analysis identifies S4 models as best predictor architecture (in combination with a shallow CNN encoder for raw time series and with no encoder for spectrogram input).}.
	\label{tab:sedf1}
	\begin{tabular}{p{0.9cm}p{0.8cm}p{0.8cm}p{0.8cm}p{0.6cm}p{0.6cm}p{0.6cm}p{0.6cm}p{0.6cm}p{0.6cm}p{0.6cm}p{0.6cm}}
		\toprule
			channels&features&encoder&predictor&\multicolumn{6}{c}{$F_1$ $(\uparrow)$}&acc. $(\uparrow)$&macro AUC $(\uparrow)$\\
		&&&&macro& W & N1 & N2 & N3 & REM\\
		\midrule 
		\multirow{11}{*}{EEG$\times$1}&\multirow{5}{*}{raw}&\textbf{CNN}&\textbf{S4}&\textbf{0.749}&\textbf{0.908}&0.421&0.841&0.811&\textbf{0.762}&\textbf{0.803}&\textbf{0.954}\\
							&	&CNN&TF&0.710&0.880&0.355&0.822&0.818&0.675&0.770&\textbf{0.954}\\
        						&	&CNN&LSTM&0.744&0.902&0.403&\textbf{0.852}&\textbf{0.821}&0.744&0.802&0.951\\
        						&	&NONE&S4&0.743&0.897&\textbf{0.454}&0.829&0.799&0.734&0.786&0.947\\
        						&	&NONE&LSTM&0.714&0.887&0.396&0.831&0.801&0.653&0.770&0.938\\
        \cline{2-12}
        
							&\multirow{6}{*}{spec}&CNN&S4&0.695&0.855&0.400&0.810&0.766&0.646&0.744&0.931\\
							&	&CNN&TF&0.723&0.882&0.386&\textbf{0.836}&0.784&0.726&0.782&0.942\\
        						&	&CNN&LSTM&0.730&0.882&0.410&0.832&0.797&0.728&0.785&0.944\\        
							& 	&\textbf{NONE}&\textbf{S4}&\textbf{0.734}&\textbf{0.884}&\textbf{0.417}&0.831&0.796&\textbf{0.742}&\textbf{0.787}&\textbf{0.949}\\
							&	&NONE&TF&0.721&0.873&0.415&0.830&0.785&0.701&0.769&0.942\\
        						&	&NONE&LSTM&0.715&0.867&0.362&0.832&\textbf{0.798}&0.717&0.775&0.937\\               	
		\midrule 
		\multirow{11}{*}{\shortstack{EEG$\times$2\\EOG$\times$1}}	&\multirow{5}{*}
							{raw}&\textbf{CNN}&\textbf{S4}&\textbf{0.780}&0.916&0.489&\textbf{0.851}&\textbf{0.822}&\textbf{0.824}&\textbf{0.823}&\textbf{0.961}\\
							&	&CNN&TF&0.745&0.913&0.439&0.811&0.802&0.760&0.793&0.950\\
        						&	&CNN&LSTM&0.767&\textbf{0.917}&0.482&0.840&0.807&0.787&0.813&0.957\\
        						&	&NONE&S4&0.776&0.910&\textbf{0.510}&0.837&0.802&0.821&0.813&0.960\\
        						&	&NONE&LSTM&0.729&0.903&0.412&0.810&0.792&0.728&0.778&0.942\\
        \cline{2-12}
        
							&\multirow{6}{*}
							{spec}&CNN&S4&0.750&0.914&0.433&0.830&0.773&0.802&0.808&0.951\\
							&	&CNN&TF&0.775&0.911&0.488&0.844&0.798&0.833&0.817&0.956\\
        						&	&CNN&LSTM&0.771&\textbf{0.920}&0.453&0.842&0.806&\textbf{0.836}&0.821&0.955\\

							& 	&\textbf{NONE}&\textbf{S4}&\textbf{0.783}&0.915&\textbf{0.510}&\textbf{0.851}&0.802&\textbf{0.836}&\textbf{0.822}&0.960\\
							&	&NONE&TF&0.777&0.915&0.494&0.850&0.799&0.827&\textbf{0.822}&\textbf{0.961}\\
        						&	&NONE&LSTM&0.775&0.913&0.479&0.846&\textbf{0.813}        						
        						
&0.825&0.820&0.959\\
        
		\bottomrule	
	\end{tabular}

\end{table}

\subsection{Model selection: multi-epoch models}
\label{sec:results_multi}

\begin{table}[!!!h]
	\centering
	\scriptsize
	\caption{Comparing different encoder and predictor architectures for \textit{15-epochs input size} trained on SEDF(full) based on validation set scores. The scores marked with an asterisk ($\ast$) represent models whose scores do not exhibit statistically significant differences (according to macro-$F_1$ score) compared to the best-performing model of the respective channel and input-modality combination. The remaining notation follows Table~\ref{tab:sedf1}. The experiments identify an S4-model-based epoch encoder (EES4) in combination with an S4-model-based predictor as the best-performing model for time series input and an S4-model-based epoch encoder (EENS4) in combination with an S4-model-based predictor as the best-performing model for spectrogram input.}
	\label{tab:15epoch}	
	
	  \begin{tabular}{p{0.9cm}p{0.8cm}p{0.8cm}p{0.8cm}p{0.6cm}p{0.6cm}p{0.6cm}p{0.6cm}p{0.6cm}p{0.6cm}p{0.6cm}p{0.6cm}}
		  \toprule
			  channels&features&encoder&predictor&\multicolumn{6}{c}{$F_1$ $(\uparrow)$}&acc. $(\uparrow)$&macro AUC $(\uparrow)$\\
		  &&&&macro& W & N1 & N2 & N3 & REM\\

		  \midrule 
		  \multirow{15}{*}{EEG$\times$1}&\multirow{6}{*}{raw}
		  		&\textbf{EES4}&\textbf{S4}&\textbf{0.801}&0.913&\textbf{0.569}&\textbf{0.868}&0.801&0.853&0.839&\textbf{0.970}\\
  
				&	&EES4&TF&0.763&0.905&0.472&0.845&0.813&0.781&0.813&0.961\\
				&	&EES4&LSTM&0.770&0.911&0.480&0.853&0.815&0.790&0.822&0.963\\
				&	&SCNN&S4&0.786&0.909&0.504&0.857&\textbf{0.819}&0.841&0.833&0.965\\					
				&	&SCNN&TF&0.789&\textbf{0.916}&0.497&0.867&0.813&0.851&\textbf{0.840}&0.967\\
				&	&SCNN&LSTM&0.781&\textbf{0.916}&0.506&0.866&0.757&\textbf{0.869}&0.834&0.966\\

		  \cline{2-12}
		  
							  &\multirow{9}{*}{spec}	
				&\textbf{EENS4}&\textbf{S4}&\textbf{0.800}&\textbf{0.919}&0.547&0.862&\textbf{0.818}&0.854&0.835&0.966\\
				&	&EENS4&TF&0.797&0.911&0.542&\textbf{0.865}&0.812&\textbf{0.861}&\textbf{0.839}&\textbf{0.968}\\
				&	&EENS4&LSTM&0.790&0.904&\textbf{0.549}&0.858&0.798&0.843&0.832&0.965\\
				&	&CNN&S4&0.764&0.889&0.472&0.848&0.787&0.821&0.810&0.956\\
				&	&CNN&TF&0.779&0.898&0.481&0.859&0.808&0.847&0.828&0.960\\		
				&	&CNN&LSTM&0.780&0.905&0.491&0.859&0.798&0.846&0.827&0.963\\				
				&	&NONE&S4&0.777&0.891&0.501&0.861&0.803&0.828&0.824&0.962\\
				&	&NONE&TF&0.741&0.876&0.438&0.840&0.783&0.767&0.789&0.950\\
				&	&NONE&LSTM&0.739&0.874&0.441&0.839&0.767&0.773&0.791&0.945\\				
										  		  
		  \midrule 
		  \multirow{15}{*}{\shortstack{EEG$\times$2\\EOG$\times$1}}	&\multirow{6}{*}{raw}
		  		&\textbf{EES4}&\textbf{S4}&\textbf{0.820}&\textbf{0.931}&\textbf{0.612}&0.862&0.798&\textbf{0.895}&\textbf{0.851}&\textbf{0.973}\\
  
				&	&EES4&TF&0.812&\textbf{0.931}&0.563&0.864&0.815&0.888&\textbf{0.851}&0.972\\
				&	&EES4&LSTM&0.803&0.925&0.538&0.862&\textbf{0.818}&0.871&0.843&0.969\\
				&	&SCNN&S4&0.787&0.919&0.515&0.840&0.809&0.854&0.830&0.964\\
				&	&SCNN&TF&0.804&0.922&0.546&\textbf{0.865}&0.817&0.872&0.844&0.969\\
				&	&SCNN&LSTM&0.795&0.922&0.537&0.860&0.786&0.870&0.840&0.968\\
		  \cline{2-12}
		  
				&\multirow{9}{*}{spec}	
				&\textbf{EENS4}&\textbf{S4}&\textbf{0.815}$^{\ast}$&
\textbf{0.931}&0.569&\textbf{0.867}&0.819&0.889&0.851&\textbf{0.970}\\
				&	&\textbf{EENS4}&\textbf{TF}&\textbf{0.815}$^{\ast}$&0.928&\textbf{0.578}&0.857&\textbf{0.827}&0.884&0.846&\textbf{0.970}\\
				&	&\textbf{EENS4}&\textbf{LSTM}&\textbf{0.815}$^{\ast}$&\textbf{0.931}&0.563&0.865&0.824&0.892&\textbf{0.853}&0.969\\
				& 	&CNN&S4&0.788&0.913&0.542&0.848&0.774&0.864&0.829&0.963\\
				&	&CNN&TF&0.788&0.909&0.506&0.854&0.801&0.872&0.832&0.966\\
							  
				&	&CNN&LSTM&0.799&0.924&0.545&0.862&0.789&0.876&0.844&0.968\\
				&	&\textbf{NONE}&\textbf{S4}&\textbf{0.815}$^{\ast}$&0.930&0.558&0.863&0.823&\textbf{0.898}&\textbf{0.853}&0.969\\
				&	&NONE&TF&0.782&0.913&0.479&0.847&0.808&0.864&0.828&0.961\\
				&	&NONE&LSTM&0.789&0.913&0.520&0.856&0.792&0.863&0.832&0.961\\	  
		  
		  \bottomrule	
	  \end{tabular}

  \end{table}

\header{Time series input} For raw time series, this study considered the best-performing encoder-predictor combination (\textit{CNN+S4}) from the previous sections as an epoch encoder (\textit{EES4}) in combination with the three different predictor architectures (\textit{S4}, \textit{TF} and \textit{LSTM}) introduced earlier. In addition, the authors investigated a strided convolutional encoder (\textit{SCNN}) in combination with these three predictors. The results of the experiments for input spanning 15-epochs are summarized in Table~\ref{tab:15epoch}.  \textit{EES4+S4} is clearly singled out as the top-performing architecture, with a significant gap to the second-best combinations  \textit{EES4+TF} in the multi-channel setting and \textit{SCNN+TF} in the single-channel setting.

\header{Spectrogram input}
As in the case of raw time series input, this study considered an epoch encoder built from the best-performing encoder-predictor combination (\textit{NONE+S4}) identified in the single-epoch experiment \textit{EENS4} in addition to the already known encoders from the single-epoch case, \text{CNN} and \textit{NONE}. As before, the results of the experiments for input spanning 15 epochs are summarized in Table~\ref{tab:15epoch}. Both in the single-channel and in the multi-channel case, the epoch encoder \textit{EENS4} is clearly singled out as the best-performing encoder architecture.  In the multi-channel case the four architectures \textit{EENS4+S4}, \textit{EENS4+TF}, \textit{EENS4+LSTM} and \textit{NONE+S4}, i.e.\ an architecture without epoch encoder, yield similar performance, while the single-channel setting is clearly dominated by \textit{EENS4+S4}.

\header{Comparative assessment}
For time series as input, the \textit{EES4} encoder represents the key component for achieving competitive performance. Overall, \textit{EES4+S4} emerges as the best-performing combination. For spectrograms as input, as the study's main focus is on identifying architectures that can generalize  to other datasets and channel configurations, this study identifies  \textit{EENS4+S4} as the architecture of choice. It is worth noting the consistent performance gaps between the best-performing models operating on time series compared to those operating on spectrograms (0.820 vs. 0.815 for multi-channel and 0.801 vs. 0.800 for single-channel). As in the case of single-epoch input models, there is a consistent improvement in multi-channel models over single-channel models.

\header{Full-epoch vs. sub-epoch encoder}
In addition to encoders for single or multiple epochs as input, experiments with sub-epoch-level encoders were conducted in this study. This study used $1/n$ of an epoch as the input unit, which corresponds to $n$ output tokens per input epoch after the encoder, and considered values for $n$ of 2, 5, and 10 for the overall best-performing model architecture (\textit{EES4+S4}). Notation-wise, this study refers to a sub-epoch encoder with $n=5$ as \textit{1/5 EES4}, when using an S4 epoch encoder. The results shown in Table~\ref{tab:subee} of the supplementary material suggest that the choice of sub-epoch level can influence the model performance. In the case of single-channel input, $n=5$ exhibits superior performance in terms of macro-$F_1$ scores compared to the single-epoch encoder ($n=1$). Even though the model performance is comparable to the standard epoch encoder in the case of multi-channel input, this study strives to identify architectures that work reliably across different situations, and therefore still selects a sub-epoch encoder ($n=5$) for the final experiments. In the case of spectrogram input, the multi-channel case favors both $n=1$ and $n=10$ but the single-channel case clearly distinguishes $n=1$. As the authors were interested in identifying an architecture that generalizes across channel configurations, the conventional epoch encoder ($n=1$) is selected for this case.

\begin{figure}[!h]
  \centering
  \begin{subfigure}[b]{0.49\textwidth}
  	\centering
  	\includegraphics[width=0.99\linewidth]{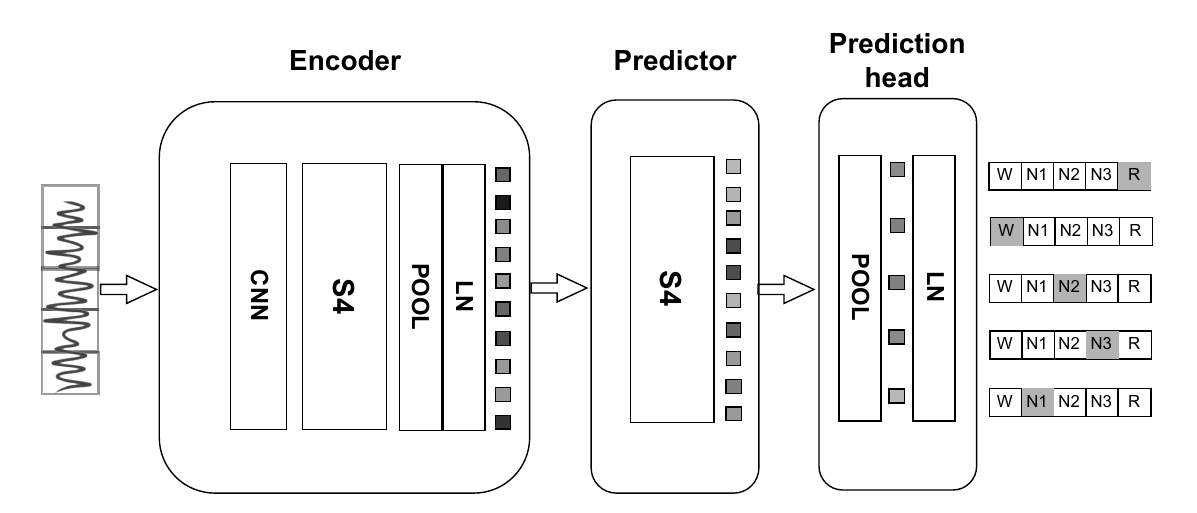}
  	\caption{S4Sleep(ts)}
  	\label{fig:s4ts}
  \end{subfigure}
  \hfill
  \begin{subfigure}[b]{0.49\textwidth}
  	\centering
  	\includegraphics[width=0.99\linewidth]{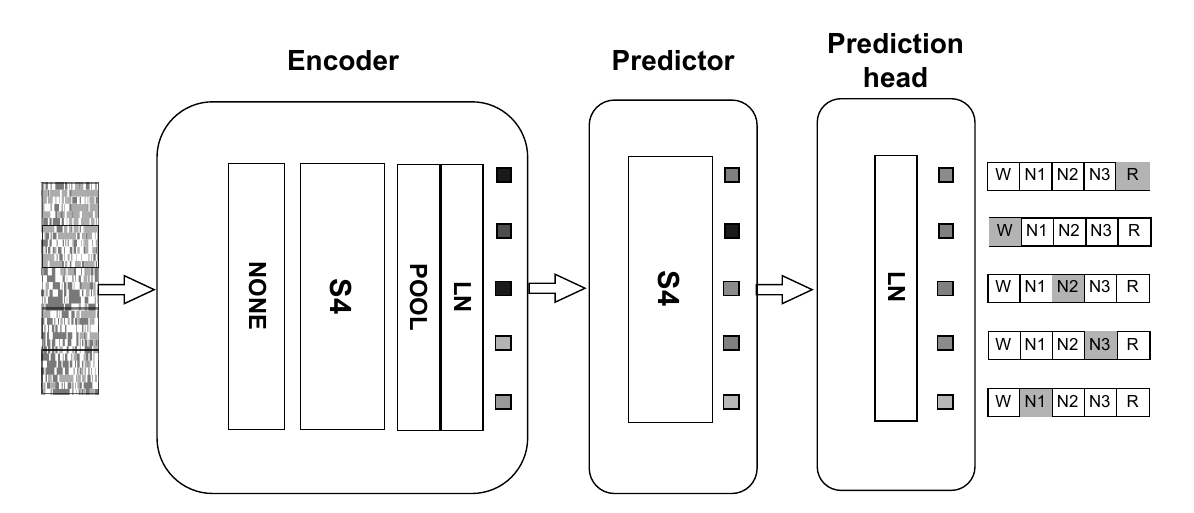}
  	\caption{S4Sleep(spec)}
  	\label{fig:s4spec}
  \end{subfigure}
  
  \caption{Schematic representation of the selected model architectures used in this work. (a) S4Sleep(ts): Raw time series as input modality; (b) S4Sleep(spec): Spectrogram as input modality.}
  \label{fig:s4sleep}
  
\end{figure}

\header{Summary}
At this point it is worthwhile to recapitulate the findings from the experiments presented in the previous paragraphs: For time series input, this study identifies \textit{1/5 EES4+S4}, i.e., an epoch encoder leveraging an S4 model operating on a sub-epoch level of 1/5 of an epoch combined with another S4 model as predictor, as the best-performing architecture and refers to it as \textit{S4Sleep(ts)}. For spectrogram input, this study identifies \textit{EENS4+S4}, i.e., again both epoch encoder and predictor leveraging an S4-model, as the best-performing architecture and refers to it as \textit{S4Sleep(spec)}. The two selected model architectures are summarized schematically in \Cref{fig:s4sleep}.

 In contrast to the majority
\subsection{Performance evaluation}
\label{sec:testseteval}
\header{Overview}
In this section, the identified architectures \textit{S4Sleep(ts)} and \textit{S4Sleep(spec)} are put to the test. First, this study evaluates the identified optimal model architectures for multi-channel and single-channel input on the SEDF(full) test set. Second, this study also presents results on the smaller SEDF20 subset, which is one of the most widely used benchmark datasets for sleep staging. Finally, as the whole analysis so far has focused on SEDF, this study aimed to demonstrate that the identified architectures also generalize to other datasets after retraining. To this end, the authors trained and evaluated the same model architectures with identical hyperparameters (with the exception of the learning rate, which was determined using the learning rate finder as described above) on the SHHS1 and MASS-SS3 datasets.

\header{Performance on SEDF(full)}
The results obtained on SEDF(full) are summarized in Table~\ref{tab:mainresult_sedf}. As a general observation, the proposed S4Sleep methods represent the best-performing methods in both the single-channel and multi-channel categories. \textit{S4Sleep(spec)} and \textit{S4Sleep(ts)} perform comparably within statistical error bars. Although literature approaches differ slightly in sample selection, splits and score evaluation methods,  assuming that the scores of these approaches are subject to similar statistical and systematic errors as the proposed models, the proposed models outperform all literature approaches in a statistically significant manner.

\header{Performance on SEDF20}
In order to allow for a comprehensive comparison with literature approaches, this study includes in Table~\ref{tab:mainresult_sedf} results for the smaller SEDF20 subset using a leave-one-patient-out cross-validation scheme, aligning with the most widely used evaluation procedure in the literature. While the performance insights from this subset may be less reliable due to its smaller size and more homogeneous patient collective, it is important to emphasize that this limitation applies equally to all studies utilizing this dataset. This approach is specifically chosen to enhance comparability with existing literature, ensuring that the findings are directly comparable to those of other researchers working under similar constraints. In the case of a single input channel, \textit{S4Sleep(ts)} is only outperformed by \textit{XSleepNet2}, while still remaining competitive with the best-performing results within error bars. For multi-channel input, \textit{S4Sleep(spec)} represents the best-performing solution but the advantage over \textit{XSleepNet2} is most likely not statistically significant. Due to the nature of the cross-validation procedure, the model will also be evaluated on samples that were placed in the validation set of the SEDF (full) dataset during the model selection phase. This might lead to a slight overestimation of the generalization capabilities of the proposed models. This makes the results on SEDF20 less reliable than those on SEDF (full) and SHHS1. The close proximity of the literature results poses a challenge for statistically significant model comparisons, raising question about  suitability of SEDF20 as a primary benchmark dataset for sleep staging algorithms.

\header{Performance on MASS-SS3}
In Table~\ref{tab:mainresult_massshhs}, the analysis is extended to include results from the MASS-SS3 subset, utilizing a 31-fold cross-validation scheme that aligns with the most commonly used evaluation scheme in the literature. Despite the previously noted limitations of this scheme, as discussed in the evaluation of the SEDF20 dataset, it provides a valuable opportunity to compare the proposed models with existing methods. The internal ranking of the two approaches, \textit{S4Sleep(ts)} and \textit{S4Sleep(spec)}, is consistent with the rankings observed in SEDF(full), where \textit{S4Sleep(spec)} outperforms \textit{S4Sleep(ts)}. In single-channel configurations, both \textit{S4Sleep(spec)} and \textit{S4Sleep(ts)} significantly outperform all other literature methods. Notably, \textit{S4Sleep(ts)} surpasses \textit{XSleepNet2}, the latter of which showed no significant difference in performance when evaluated on the SEDF20 single-channel configuration. For multi-channel inputs, \textit{S4Sleep(spec)} continues to significantly outperform all other literature results, while \textit{S4Sleep(ts)} matches the performance of other top-performing models and only slightly underperforms GraphSleepNet.

\header{Performance on SHHS1}
The extensive SHHS1 dataset, previously unseen during the model selection procedure, serves as an objective evaluation of the robustness of the achieved findings. Although existing literature often lacks detailed disclosure regarding dataset divisions, the substantial volume of the SHHS1 dataset significantly diminishes the impact of this variability on the comparability of results. The results demonstrate minimal variation in performance, underscoring the robustness and reliability of the models. To foster greater transparency and reproducibility in research, the authors have provided details on their dataset split.
First of all, it is noticeable from results presented in Table~\ref{tab:mainresult_massshhs} for SHHS1 that both \textit{S4Sleep(ts)} and \textit{S4Sleep(spec)} models outperform the current state-of-the-art performance for both considered channel configurations. In contrast to the majority of literature approaches, this study provides statistical and systematic uncertainty estimates. Both are on the order of 0.001 and therefore comparably small, which underscores the robustness of the results. Unfortunately, no error estimates for the literature results are available. However, under the assumption that the literature results are subject to comparable statistical error estimates as the proposed approaches, \textit{S4Sleep(ts)} would not overlap with the best  \textit{L-SeqSleepNet}-performance, which represents a significant improvement in performance in the single-channel
case. In the multi-channel case, the performance of \textit{S4Sleep(spec)}, with a macro-$F_1$ score of 0.835, is significantly better compared to \textit{XSleepNet2}, which achieved a macro-$F_1$ score of 0.823. This difference is deemed significant, assuming comparable statistical uncertainties in the literature results. At this point one has to bear in mind that there is presently no generally agreed train-test split available for SHHS1, a shortcoming that this study aims to address by providing train-test splits as part of the code repository \cite{repo} accompanying this submission. On the one hand, this means that scores are, strictly speaking, not directly comparable. On the other hand, SHHS1 is already a large-scale dataset, where score fluctuations due to the size of the test set are not expected to play a major role. This argument is supported by the comparably small statistical uncertainties demonstrated in the analysis. As a final remark, it is worth stressing that the exceptionally strong performance of both \textit{S4Sleep} variants without further hyperparameter adjustments is a very good sign for the generalization capabilities of the proposed architecture.

\begin{table}[!h]
	\centering
	\scriptsize
	\caption{Performance of selected models on SEDF(full) and SEDF20. Best-performing models according to different evaluation metrics are highlighted in boldface. For SEDF(full), both systematic uncertainties, assessed via three training runs (first entry), and the statistical uncertainty due to the composition of the test set, assessed via bootstrapping (second entry), are reported separately. For SEDF20, only systematic uncertainties assessed via three training runs are reported. References marked with a ${}^\ast$ symbol indicate the most recent works, published in 2023 or later.} 
	\label{tab:mainresult_sedf}
		
	\begin{tabular}{p{0.21cm}p{3.6cm}p{0.5cm}p{1.5cm}p{0.42cm}p{0.42cm}p{0.42cm}p{0.42cm}p{0.42cm}p{0.43cm}p{0.43cm}}
	
		\toprule
			& model & channel	& \multicolumn{6}{c}{$F_1$ $(\uparrow)$} & acc. $(\uparrow)$ & macro AUC $(\uparrow)$ \\
		&&&macro& W & N1 & N2 & N3 & REM&& \\
		
		\midrule 
		
		\multirow{15}{*}{\rotatebox[origin=c]{90}{SEDF(full)}} 
		& XSleepNet1\textsuperscript{c,k}\cite{phan2021xsleepnet} & single &  0.778 & - & - & - & - & - & 0.836 & - \\
		& XSleepNet2\textsuperscript{c,k}\cite{phan2021xsleepnet} & single & 0.779 & - & - & - & - & - & 0.840 & - \\
		& SleepTransformer\textsuperscript{c,k}\cite{phan2022sleeptransformer} & single &  0.743 & 0.917 & 0.404 & 0.843 & 0.779 & 0.772 & 0.814 & - \\ 

		& S4Sleep(ts)\textsuperscript{a,n}  & single &  0.796$\pm$0.004 $\pm$0.004 & \textbf{0.925} & 0.503 & 0.854& \textbf{0.839} & 0.855 & 0.839 & 0.958 \\
		& \textbf{S4Sleep(spec)}\textsuperscript{a,n} & single & \textbf{0.807$\pm$0.008} \textbf{$\pm$0.003}  & 0.923 & \textbf{0.558} & \textbf{0.861}  & 0.821 & \textbf{0.877} & \textbf{0.844} & \textbf{0.964} \\

		\cline{2-11} 
		
		& XSleepNet1\textsuperscript{c,k}\cite{phan2021xsleepnet} & multi &  0.774 & - & - & - & - & - & 0.840 & - \\
		& XSleepNet2\textsuperscript{c,k}\cite{phan2021xsleepnet} & multi & 0.777 & - & - & - & - & - & 0.840 & - \\
		& RobustSleepNet\textsuperscript{c,j}\cite{guillot2021robustsleepnet} & multi &  0.763 & - & - & - & - & - & - & - \\
		& RobustSleepNet\textsuperscript{a,j}\cite{guillot2021robustsleepnet} & multi &  0.778 & - & - & - & - & - & - & - \\
		& U-Sleep\textsuperscript{c,q}\cite{Perslev2021} & multi &  0.79 & \textbf{0.93} & 0.57 & 0.86 & 0.71 & 0.88 & - & - \\
		& SAGSleepNet\textsuperscript{a,l,$\ast$}\cite{jin2023sagsleepnet} & multi &  0.744$\pm$0.052 & - & - & - & - & - & 0.807 & 0.951 \\		
		& S4Sleep(ts)\textsuperscript{a,n} & multi &  0.804$\pm$0.002 $\pm$0.003 & 0.925 & 0.568 & 0.847 & \textbf{0.826} & 0.863 & 0.841 & 0.961\\
		& \textbf{S4Sleep(spec)}\textsuperscript{a, n} & multi 	& \textbf{0.817$\pm$0.007} \textbf{$\pm$0.003}  & 0.928& \textbf{0.571} & \textbf{0.867} & 0.814 &  \textbf{0.906} & \textbf{0.853} & \textbf{0.971} \\
		
		\midrule 

		\multirow{15}{*}{\rotatebox[origin=c]{90}{SEDF20}} 
		& DeepSleepNet\textsuperscript{b,l}\cite{supratak2017deepsleepnet} & single &  0.769 & 0.847 & 0.466  & 0.859 & 0.848 & 0.824 & 0.820 & - \\	
		& SeqSleepNet\textsuperscript{b,l}\cite{phan2019seqsleepnet} & single &  0.786$\pm$0.002 & 0.912 & 0.447 & 0.880 & 0.862 & 0.830 & 0.856 & - \\ 
		& IITNet\textsuperscript{b,l}\cite{seo2020intra} & single &  0.776 & 0.877 & 0.434 & \textbf{0.898} & 0.848 & 0.824 & 0.820 & - \\
		& XSleepNet1\textsuperscript{b,l}\cite{phan2021xsleepnet}\cite{phan2023lseqsleepnet} & single &  0.800 & 0.913 & 0.495 & 0.880 & \textbf{0.869} & 0.842 & 0.860 & - \\
		& \textbf{XSleepNet2}\textsuperscript{b,l} \cite{phan2021xsleepnet}\cite{phan2023lseqsleepnet} & single & \textbf{0.806} & \textbf{0.922} & 0.518 & 0.880 & 0.868 & 0.839 & 0.863 & - \\
		& An et al 2022\textsuperscript{d,l}\cite{an2022amplitude}  & single &  0.801 & -   & -  &  - & -  & -  &  \textbf{0.931} & -\\		
		& L-SeqSleepNet\textsuperscript{b,l,$\ast$}\cite{phan2023lseqsleepnet} & single &  0.793$\pm$0.004 & 0.916 & 0.453 & 0.885 & 0.862 & 0.852 & 0.863 & - \\	

		& S4Sleep(ts)\textsuperscript{b,l}  & single &  0.804$\pm$0.004 & 0.906      & \textbf{0.524}  &  0.874 & 0.862  & \textbf{0.853}  &  0.876 &\textbf{0.971}\\
		& S4Sleep(spec)\textsuperscript{b,l} & single & 0.787$\pm$0.004  & 0.892 & 0.493 & 0.865 &0.842 & 0.847 & 0.849 & 0.970 \\

		\cline{2-11} 
		
		& XSleepNet1\textsuperscript{b,l}\cite{phan2021xsleepnet} & multi &  0.798 & - & - & - & - & - & 0.852 & - \\
		& XSleepNet2\textsuperscript{b,l}\cite{phan2021xsleepnet} & multi & 0.809 & - & - & - & - & - & 0.864 & - \\ 
		& RobustSleepNet\textsuperscript{b,j}\cite{guillot2021robustsleepnet} & multi &  0.791 & - & - & - & - & - & - & - \\
		& ProductGraphSleepNet\textsuperscript{b,l,$\ast$}\cite{einizade2023productgraphsleepnet} & multi &  0.774 & 0.886 & 0.426 & 0.874 & 0.847 & 0.834 & 0.838 & - \\
		& S4Sleep(ts)\textsuperscript{b,l} & multi & 0.805$\pm$0.004  & 0.894 & \textbf{0.521} & 0.876 & \textbf{0.858} & 0.873 & 0.860 & 0.971 \\
		& \textbf{S4Sleep(spec)}\textsuperscript{b,l}  & multi &  \textbf{0.810$\pm$0.007} & \textbf{0.922}  &   0.519 & \textbf{0.879}  & 0.857  & \textbf{0.877}  & \textbf{0.866}  &  \textbf{0.979} \\		
							       
		\bottomrule	
	
	\end{tabular}

	\parbox{\linewidth}{\tiny a: based on full SEDF. b: based on SEDF20. c: based on SEDF-SC. d: based on recordings from 20 selected subjects in the full  SEDF. j: 5-fold cross-validation. k: 10-fold cross-validation. l: leave-one-subject-out cross-validation. n: 8:1:1 training-validation-test holdout set evaluation. q: tested on 23 recordings from SEDF-SC.}	
    
\captionsetup{justification=centering, singlelinecheck=false, position=top}
\end{table}

\begin{table}[!h]
	\centering
	\scriptsize

	\caption{Performance of selected models on MASS-SS3 and SHHS1 (test set scores). Best-performing models according to different evaluation metrics are highlighted in boldface. For MASS-SS3, only systematic uncertainties assessed via multiple training runs are reported. For SHHS1, both systematic uncertainties assessed via three training runs (first entry) and the statistical uncertainty due to the composition of the test set assessed via bootstrapping (second entry) are reported. References marked with a ${}^\ast$ symbol indicate the most recent works, published in 2023 or later.}
	\label{tab:mainresult_massshhs}
		
	\begin{tabular}{p{0.23cm}p{3.1cm}p{0.5cm}p{2.4cm}p{0.42cm}p{0.42cm}p{0.42cm}p{0.42cm}p{0.42cm}p{0.43cm}p{0.43cm}}
	
		\toprule
			 & model & channel	& \multicolumn{6}{c}{$F_1$ $(\uparrow)$} & acc. $(\uparrow)$ & macro AUC $(\uparrow)$ \\
		&&&macro& W & N1 & N2 & N3 & REM&& \\						    

		\midrule 
		\multirow{13}{*}{\rotatebox[origin=c]{90}{MASS}} 
		& DeepSleepNet\textsuperscript{a,k}\cite{supratak2017deepsleepnet} & single &  0.817 & 0.873 & 0.598  & 0.903 & 0.815 & 0.893 & 0.862 & - \\			
		& IITNet\textsuperscript{a,k}\cite{seo2020intra} & single &  0.800 & 0.852 & 0.518 & \textbf{0.914} & \textbf{0.869} & 0.845 & 0.862 & - \\
		& XSleepNet1\textsuperscript{b,j}\cite{phan2021xsleepnet} & single &  0.806 & - & - & - & - & - & 0.851 & - \\  
		& XSleepNet2\textsuperscript{b,j}\cite{phan2021xsleepnet} & single &  0.806 & - & - & - & - & - & 0.852 & - \\
		& S4Sleep(ts)\textsuperscript{a,k}  & single &  0.821$\pm$0.001 & 0.876 & 0.608 & 0.907 & 0.815 &  \textbf{0.900} & 0.873 & 0.978 \\ 
		& \textbf{S4Sleep(spec)}\textsuperscript{a,k} & single & \textbf{0.824$\pm$0.001} & \textbf{0.877} & \textbf{0.619} & 0.912 & 0.814 &0.898 &  \textbf{0.877} & \textbf{0.979} \\ 

		\cline{2-11}
		& SeqSleepNet\textsuperscript{b,j}\cite{phan2019seqsleepnet} & multi &  0.833 & 0.890 & 0.597 & 0.909 & 0.802 &  \textbf{0.935} & 0.871 & - \\ 
		& XSleepNet1\textsuperscript{b,j}\cite{phan2021xsleepnet} & multi  & 0.837 & - & - & - & - & - & 0.875 & - \\  
		& XSleepNet2\textsuperscript{b,j}\cite{phan2021xsleepnet} & multi  & 0.838 & - & - & - & - & - & 0.876 & - \\
		& RobustSleepNet\textsuperscript{a,i}\cite{guillot2021robustsleepnet} & multi &  0.822 & - & - & - & - & - & - & - \\
		& GraphSleepNet\textsuperscript{a,k,$\ast$}\cite{jin2023sagsleepnet} & multi &  0.841 &\textbf{0.913} & 0.603 & \textbf{0.921} &\textbf{0.851} & 0.919 &  \textbf{0.889} & - \\
		& S4Sleep(ts)\textsuperscript{a,k} & multi & 0.837$\pm$0.002 & 0.887 & 0.631 & 0.920 & 0.842 & 0.906 & 0.887 & 0.982 \\ 
		& \textbf{S4Sleep(spec)}\textsuperscript{a,k} & multi & \textbf{0.843$\pm$0.001} & 0.907 & \textbf{0.670} & 0.917 & 0.806 & 0.911 & \textbf{0.889} & \textbf{0.984} \\

		\midrule 
		\multirow{13}{*}{\rotatebox[origin=c]{90}{SHHS1}} 		
		& SeqSleepNet\textsuperscript{f,p}\cite{phan2019seqsleepnet} & single &  0.802 & 0.918 & 0.491 & 0.882 & 0.835 & 0.882 & 0.872 & - \\ 
		& IITNet\textsuperscript{f,o}\cite{seo2020intra} & single &  0.798 & 0.901 & 0.481 & 0.884 & 0.852 & 0.872 & 0.867 & - \\
		& XSleepNet1\textsuperscript{f,p}\cite{phan2021xsleepnet} & single &  0.807 & 0.916 & 0.514 & 0.885 & 0.850 & 0.884 & 0.876 & - \\  
		& XSleepNet2\textsuperscript{f,p}\cite{phan2021xsleepnet} & single &  0.810 & 0.920 & 0.499 & 0.883 & 0.850 & 0.882 & 0.875 & - \\
		& SleepTransformer\textsuperscript{f,p}\cite{phan2022sleeptransformer} & single &  0.801 & 0.922 & 0.461 & 0.883 & 0.852 & 0.886 & 0.877 & - \\
		& L-SeqSleepNet\textsuperscript{f,p,$\ast$}\cite{phan2023lseqsleepnet} & single &  0.814 & 0.931 & 0.511 & \textbf{0.890} & 0.849 & 0.898 & \textbf{0.884} & - \\   
		& S4Sleep(ts)\textsuperscript{f,p}  & single &  0.817$\pm$0.001$\pm$0.001 & 0.931 & 0.526 & 0.888 & 0.848 & 0.895 & 0.882 & \textbf{0.977} \\ 
		& \textbf{S4Sleep(spec)}\textsuperscript{f,p} & single & \textbf{0.819$\pm$0.002} \textbf{$\pm$0.0005}  & \textbf{0.932} & \textbf{0.527}&0.885  & \textbf{0.853}& \textbf{0.899} & 0.883 & \textbf{0.977} \\ 

		\cline{2-11}

		& XSleepNet1\textsuperscript{f,p}\cite{phan2021xsleepnet} & multi  & 0.822 & - & - & - & - & - & 0.891 & - \\  
		& XSleepNet2\textsuperscript{f,p}\cite{phan2021xsleepnet} & multi  & 0.823 & - & - & - & - & - & 0.891 & - \\
		& RobustSleepNet\textsuperscript{g,i}\cite{guillot2021robustsleepnet} & multi &  0.792 & - & - & - & - & - & - & - \\
		& U-Sleep\textsuperscript{h,r}\cite{Perslev2021} & multi  & 0.80 & 0.93 & 0.51 & 0.87 & 0.76 & 0.92 & - & - \\
		& Pei et al 2024\textsuperscript{e,m,$\ast$}\cite{pei2024automatic} & multi &  0.693 & 0.938 & 0.271 & 0.795 & 0.641 & 0.822 & 0.824 & - \\
		& S4Sleep(ts)\textsuperscript{f,p}  & multi & 0.827$\pm$0.001$\pm$0.002 & 0.938 & 0.536 & 0.894 & 0.849 & 0.919 & 0.890 & 0.980 \\ 
		& \textbf{S4Sleep(spec)}\textsuperscript{f,p} & multi & \textbf{0.835$\pm$0.001} \textbf{$\pm$0.001} & \textbf{0.944}  & \textbf{0.555} & \textbf{0.896}&\textbf{0.857} &\textbf{0.923} &\textbf{0.896}  &\textbf{0.982} \\        					    
               
		\bottomrule	
	
	\end{tabular}

	\parbox{\linewidth}{\tiny a: based on MASS-SS3. b: based on MASS SS1-SS5. e: based on recordings from 100 selected subjects in the SHHS visit 1. f: based on SHHS Visit 1. g: based on SHHS Visit 2. h: based on the full SHHS. i: 5-fold cross-validation. j: 20-fold cross-validation. k: 31-fold cross-validation.  m: 7:2:1 training-validation-test holdout set evaluation. o: 5:2:3 training-validation-test holdout set evaluation. p: 7:3 training-test holdout set evaluation. r: tested on 140 recordings from the full SHHS.}	
    
\captionsetup{justification=centering, singlelinecheck=false, position=top}
\end{table}

\section{Discussion}
\label{sec:discussion}
\header{Search for model architectures}
This paper demonstrates the technical feasibility of identifying optimal model architectures from a relatively large class of encoder-predictor models in a structured fashion while also keeping track of the uncertainties associated with the approach. The authors envision that this procedure could serve as a blueprint for other long-term time series classification tasks in other domains.

\header{Time-series-based vs. spectrogram-based models}
In multi-channel cases,  \textit{S4Sleep(spec)} consistently outperforms  \textit{S4Sleep(ts)} across all evaluated datasets. An important question arises: how can time-series-based models be developed to match or surpass the performance of spectrogram-based methods in these instances? Similarly, in single-channel cases,  \textit{S4Sleep(spec)} generally dominates, though the results are often close and sometimes overlap within error margins. The SEDF20 dataset is a notable exception, where  \textit{S4Sleep(ts)} clearly outperforms  \textit{S4Sleep(spec)}. This dataset is unique as it is the smallest and oldest among those considered, likely exhibiting the lowest signal quality. It can be hypothesized that this might lead to robustness challenges for  \textit{S4Sleep(spec)}, while  \textit{S4Sleep(ts)} may be more adaptable in selecting robust input features. Supporting this hypothesis, the top performers among the nine models referenced in Table~\ref{tab:mainresult_sedf}—specifically XSleepNet2, \textit{S4Sleep(ts)}, An et al. (2022), and XSleepNet1—that excel on the single-channel SEDF20 dataset all incorporate time series inputs.

\header{Single-epoch vs. multi-epoch models}
This study also used the established methodology to  assess the statistical significance of using multiple instead of a single epoch as input. To this end, for a given channel configuration and input modality, the authors compared the best-performing single-epoch model as identified in Table~\ref{tab:sedf1} and the best-performing multi-epoch model as identified by Table~\ref{tab:15epoch}. In all cases, the study  demonstrated significant performance differences between single- and multi-epoch models at a predefined threshold of 60\% of pairwise tests following the methodology described in \ref{app:train_and_eval}, as corroborated by the individual single-epoch results in Table~\ref{tab:mainresult_singleepoch} in the supplementary material. These results reiterate the empirical findings of \cite{phan2019seqsleepnet} and emphasize the significant advantage of jointly predicting multiple sleep stages for multiple input epochs at once, which enables the incorporation of more contextual information into each single-epoch prediction.

\header{Long-range interactions}
It is worth pointing out that \textit{S4Sleep(ts)}, alongside \cite{an2022amplitude}, is one of the few approaches that demonstrates the competitiveness of raw time series as a standalone input representation for sleep staging. In particular, \textit{S4Sleep(ts)} achieved a score that is competitive or even superior to L-SeqSleepNet both on SHHS1 and SEDF20 without explicitly leveraging long-range correlations across hundreds of sleep epochs, which the refereed study identifies as a central factor for the performance improvement over prior state-of-the-art. A first exploratory study \cite{wang2024assessing} building on the \textit{S4Sleep(ts)} architecture proposed in this work, showed no performance improvements upon increasing the input size even after gradual upscaling, calling into question the diagnostic relevance of long range interactions across hundreds of input epochs.

\header{Comparison to literature approaches}
The strong results achieved by the proposed methods should be seen as an indication of the effectiveness of encoder-predictor models for the purpose of sleep staging and, most likely, long time series annotation tasks in general. This could also be seen as a recommendation for adopting greater architectural simplicity as many of the competing approaches that fall into the encoder-predictor category, see \Cref{tab:architectures}, rely on custom model components such as specific encoder or pooling layers. Apart from the recently proposed S4-layer, which can serve as a drop-in replacement for recurrent or transformer layers, this study only leveraged standard model components.

\header{Model complexity}
Finally, it is worth stressing that all performance comparisons should be set into perspective with the complexity of the underlying model. Even though the authors acknowledge that the number of model parameters is an imperfect proxy for model complexity, it is the most widely reported in the literature. In Table~\ref{tab:model-size-training-time} of the supplementary material, this study compares parameter counts of the proposed models to those of models used in the literature. The proposed models achieve parameter counts around $5\times 10^6$ and therefore fall in a similar range as competing approaches. If parameter counts are accepted as a proxy for model complexity, the strong performance of the proposed models cannot be attributed to an excessively complex model.

\header{Limitations and future work}
This work has several limitations. First, the evaluation focuses on predictive accuracy and model complexity, while excluding other important quality dimensions, such as robustness to input noise and distribution shifts, and model explainability. The authors believe that investigating these quality dimensions is crucial for clinical applications and merit dedicated studies. Second, this study currently lacks a deeper understanding of why the two input representations seem to favor different encoder choices (\textit{CNN} for raw time series and \textit{NONE} for spectrograms). On a related note, the authors believe that a more elaborate handling of multiple input channels could enable raw-time-series-based approaches to reach the same level of performance as spectrogram-based approaches. Third, this study deliberately focused on comparing models trained from scratch. The authors consider the investigation of different training paradigms, such as supervised or self-supervised pretraining schemes for the identified architectures, as a very promising direction for future research. Finally, the authors interpret the fact that both model variants show strong performance on SHHS1 for both the single- and multi-channel input representations without any further hyperparameter adjustments as a sign of the generalization capabilities of the proposed model architectures. Therefore, they consider this a very promising step to also explore these models in the context of other data modalities with physiological time series, such as long-term ECG/PPG-based arrhythmia classification, or even for time series classification tasks beyond physiological time series, such as audio classification or other domains. This argument is supported by a recent independent study \cite{bhirangi2024hierarchicalstatespacemodels}, which identified very similar architectures in the context of robotic sensor data.

\section{Summary and conclusion}
\label{sec:summary}
This study addresses the problem of automatic sleep staging from polysomnography, which represents a time series classification task where sleep stages have to be predicted per epoch of 30 seconds. Based on the specific task, the authors devised a systematic procedure to identify optimal model architectures for long time series annotation tasks within the search space of encoder-predictor architectures, which encompasses the majority of literature approaches. This process led to the identification of two optimal model architectures, \textit{S4Sleep(ts)} and \textit{S4Sleep(spec)}, for both raw time series and spectrograms as input. Both architectures leverage structured state space models as essential components to extract discriminative representations from the input, albeit in an input-modality specific manner. On the full SEDF, MASS-SS3 and SHHS1  datasets, the proposed models outperform all literature approaches in a statistically significant manner, both for single-channel and multi-channel input.  \textit{S4Sleep(spec)} outperformed  \textit{S4Sleep(ts)} in seven out of eight benchmarking scenarios, highlighting  the discriminative power of spectrograms as an input representation for automatic sleep staging.    The code underlying the experiments presented in this study is publicly available \cite{repo}.

\section*{Acknowledgments}

The authors thank  Dr. Insa Wolf and Fabian Radtke for valuable discussions.

\clearpage
\bibliographystyle{elsarticle-num} 
\bibliography{bibfile}



\clearpage
\appendix
\setcounter{page}{1}
\section{Model architectures}
\label{app:architectures}
\subsection{Single-epoch prediction models}
\label{app:sing-ep}

\header{Encoder architectures} For a single-epoch time series input, this study utilized an encoder architecture that comprises two one-dimensional convolutional layers. Each layer consists of 128 features, a kernel size of 3, a stride of 2, corresponding to a moderate downsampling factor of 4, interleaved by ReLU activation functions. In the case of spectrogram input, the authors reshape the input to join frequency and channel axes, which subsequently allows for further processing using one-dimensional convolutions. In this case, the authors employ a similar encoder structure with two 1D convolutional layers. Here, they use 128 features, a kernel size of 3 and a stride of 1 interleaved by GeLU activation functions. This particular choice was inspired by the encoder used in state-of-the-art speech recognition models operating on spectrograms as input data \cite{radford2023robust} and reduces to the former in the case of a single input channel. To simplify the notation, this study refers to both encoders as \textit{CNN} encoder, denoting the the respective convolutional encoders for either time series or spectrogram data. The authors also conducted experiments with architectures without explicit encoders (\textit{NONE}), where the input is passed directly to the respective predictor models. In the case of time series input, the input sequence is just processed through a linear layer mapping from the number of input channels to the appropriate model dimension of the predictor model. In the case of spectrogram input, the authors first reshape channel and frequency axes into a common axis and then proceed with a linear layer as in the case of time series input. 

\header{Predictor architectures} This study considers three different architectures: A two-layer, bidirectional LSTM model \cite{hochreiter1997long}, a four-layer transformer model \cite{vaswani2017attention}, or a four-layer S4 model \cite{Gu2021EfficientlyML}. To ensure that the total number of trainable parameters and hence the model complexity remains comparable, the authors set the model dimensions to 512, 256, and 512 for LSTM, transformers, and S4, respectively. If the number of output features of the encoder does not match the predictor's internal feature dimension (for example in the case of \textit{NONE} encoders), the predictor is preceded by a linear layer to adjust the input features for the predictor model. The dropout rates stand at 0.1 for the transformer models and 0.2 for the S4 models. Each transformer layer features 8 heads, while the state dimension of the S4 model is set to 64. The authors omitted the transformer predictor coupled with the \textit{NONE} encoder from the evaluation, attributing to the substantially increased computational time associated with this particular configuration.

\header{Prediction head architectures} The prediction head architecture just involves a global average pooling followed by a single linear layer. The authors refrain from testing more complex pooling options such as attention-based pooling layer in order to keep the complexity of the search space manageable.

\subsection{Multi-epoch prediction models}
\label{app:multi-ep}
For multi-epoch time series input, the study considers different strategies.

\header{Epoch encoders} The first option is to use entire single-epoch models, which have shown superior performance, as encoders. This study refers to such encoders as \textit{epoch encoders}. The authors distinguish different epoch encoders based on the respective encoder-predictor combinations they use. The authors designate the respective epoch encoders based on their encoder-predictor combinations as \textit{EES4} for \textit{CNN+S4}, and \textit{EENS4} for \textit{NONE+S4}.

\header{Strided convolutions for raw time series input} The second option to reach a heavily temporally downsampled intermediate representation that is appropriate for further processing is through the use of strided convolutions. This approach is inspired by encoders used in speech recognition models operating on raw waveform data such as Wave2vec \cite{baevski2020wav2vec}. Here, the study use four 1D convolutional layers each with feature sizes 512, kernel sizes of 9-9-3-3, and stride sizes of 5-5-2-2, resulting in a temporal downsampling factor of 100, interleaved with ReLU activations. This study refers to this encoder as \textit{SCNN} (for strided CNN) encoder. 

\header{Single-epoch encoders for spectrogram input} For multi-epoch spectrogram input, one has the option to utilize the CNN encoder (\textit{CNN}) described in the paragraph on single-epoch encoder. In this case, there is no need for a strong temporal downsampling as the spectrograms by themselves already come at a 100 times smaller temporal resolution compared to the original time series input. In this sense, the simple CNN encoder for spectrograms can be regarded as analogue of the strided CNN encoder for raw time series. In addition, this study also investigates a second choice from the single-epoch encoder model, namely the \textit{NONE} encoder, i.e., omitting the encoder altogether.

\header{Predictor architectures} The three predictor options coincide with the predictor architectures described in the single-epoch case in Section \ref{app:sing-ep}.

\header{Prediction head architectures} The prediction head architecture consists of a single linear layer. Depending on the used encoder choice (i.e., for sub-epoch encoder, \textit{SCNN}, \textit{CNN}, and \textit{NONE} encoders) it is preceded by a local pooling layer with appropriate kernel size to reduce the temporal resolution of the output to one output token per epoch.

\subsection{Parameter counts}

Table~\ref{tab:model-size-training-time} summarizes parameter counts for the proposed models and selected models from the literature.
	
\begin{table}[h]
	\caption{Parameter counts for literature approaches and models proposed in this work.}
	\label{tab:model-size-training-time}
	\scriptsize
	\centering
	\begin{tabular}{p{4cm} p{3cm}}
	\hline
	\textbf{Model} & \textbf{\#parameters} \\ \hline 
	DeepSleepNet \cite{supratak2017deepsleepnet} & \(2.30 \times 10^7\) \\
	SeqSleepNet \cite{phan2019seqsleepnet} & \(1.64 \times 10^6\) \\
	U-Sleep \cite{Perslev2021} & \(3.10 \times 10^6\) \\
	RobustSleepNet \cite{guillot2021robustsleepnet} & \(1.80 \times 10^5\) \\
	XSleepNet \cite{phan2021xsleepnet} & \(5.74 \times 10^6\) \\
	SleepTransformer \cite{phan2022sleeptransformer} & \(3.70 \times 10^6\) \\
	L-SeqSleepNet \cite{phan2023lseqsleepnet} & \(6.30 \times 10^5\) \\
	\textit{S4Sleep(ts)} & \(4.90 \times 10^6\) \\
	\textit{S4Sleep(spec)} & \(4.80 \times 10^6\) \\
	\hline
	\end{tabular}

\end{table}

Table~\ref{tab:training-time} summarizes training times for the proposed models. The training was performed on a single NVIDIA A100 GPU with 80GB of memory.

\begin{table}[h]
\caption{Approximate training times for the S4Sleep models on the four considered datasets.}
\label{tab:training-time}
\centering
\scriptsize
\begin{tabular}{ccccc}
\hline
\multirow{2}{*}{\textbf{Model}} & \multirow{2}{*}{\textbf{Dataset}} & \multirow{2}{*}{\textbf{Epoch Count}} & \multicolumn{2}{c}{\textbf{Running Time (Hours)}} \\ \cline{4-5} 
                                &                                   &                                       & \textbf{Singel-channel}  & \textbf{Multi-channel} \\ \hline
\multirow{4}{*}{S4Sleep(ts)}    & SEDF(full)                        & 50 epochs                             & 4.10                     & 5.21                   \\
                                & SEDF20                            & 1 fold 50 epochs                      & 0.52                     & 0.74                   \\
                                & MASS-SS3                          & 1 fold 50 epochs                      & 0.73                     & 0.86                   \\
                                & SHHS1                             & 30 epochs                             & 59                       & 66                     \\ \hline
\multirow{4}{*}{S4Sleep(spec)}  & SEDF(full)                        & 50 epochs                             & 0.70                     & 0.81                   \\
                                & SEDF20                            & 1 fold 50 epochs                      & 0.07                     & 0.08                   \\
                                & MASS-SS3                          & 1 fold 50 epochs                      & 0.17                     & 0.30                   \\
                                & SHHS1                             & 30 epochs                             & 41                       & 48                     \\ \hline
\end{tabular}
\end{table}
	
\section{Datasets and Preprocessing}
\label{app:datasets}

\header{Sleep EDF (SEDF)} The SEDF database \cite{kemp2000analysis, goldberger2000physiobank} comprises 197 sleep recordings, sourced from two specific subsets: the Sleep Cassette (SC), which includes 153 recordings from 78 healthy individuals, and the Sleep Telemetry (ST), featuring 44 recordings from 22 patients experiencing mild difficulties in falling asleep. All recordings include Fpz-Cz and Pz-Oz EEG channels, as well as EOG channels, all sampled at 100 Hz.
The recordings are manually scored into wake stage (W), NREM stages (N1-N4), rapid-eye-movement stage (REM), MOVEMENT and UNKNOWN according to the R\&K guidelines. Following common preprocessing practices in the literature, the authors merged N3 and N4 (into N3), and excluded data segments labeled as MOVEMENT and UNKNOWN. The study conducted experiments on multi-channel data containing two EEG channels (Fpz-Cz and Pz-Oz) and one EOG channel (horizontal), as well as on single-channel data consisting only of the Fpz-Cz EEG channel. This study covers both single- and multi-channel inputs to compare as broadly as possible to available literature results. The authors performed a random split of the recordings, allocating 80\% of the recordings for training, 10\% for validation and 10\% for testing. As no standard splits are available in the literature,  the splits are released for later reproducibility \cite{repo}. A selective subset of the Sleep Cassette (SC) subset, designated as SEDF20 in this work, comprises 39 polysomnography (PSG) recordings from 20 subjects (10 males and 10 females) aged between 25 and 34, which is frequently utilized in studies focusing on automatic sleep staging. To enable a meaningful comparison of the outcomes of this study with existing literature, the authors also conducted experiments with the final selected models on this specific subset. A 20-fold cross-validation was conducted, always using 17 subjects for training, 2 for validation, and 1 for evaluation.

\header{Montreal Archive of Sleep Studies (MASS)} The MASS dataset is a comprehensive collection of PSG recordings spanning 200 nights, organized into five subsets (SS1-SS5) \cite{OReilly2014}. Due to variations in available input channels and annotation standards across subsets, the study followed the established convention in the literature and focused exclusively on the third subset, MASS-SS3. This subset includes 62 full-night PSG recordings, comprising 20 EEG channels, 3 EMG channels, 2 EOC channels, and 1 ECG channel. Sleep stages were manually scored following AASM guidelines into W, NREM stages (N1-N3), REM, and an "UNKNOWN" category. For channel configuration, the study adhered to commonly used setups for consistency with prior work. Single-channel analyses utilized the F4-EOG configuration as described in \cite{supratak2017deepsleepnet}, while multi-channel analyses incorporated all available PSG channels following \cite{jin2023sagsleepnet}. To ensure comparability with other studies using MASS-SS3, the research adopted the standard 31-fold cross-validation scheme for model evaluation, thereby providing a robust and reproducible benchmark.

\header{Sleep Heart Health Study (SHHS)} The SHHS dataset includes polysomnography data collected from participants across two overnight visits (Visit 1 and Visit 2), with several intervening years for follow-up \cite{ross2019probabilistic, quan1997sleep}. The dataset is available for download and usage through the National Sleep Research Resource (NSRR) platform\cite{zhang2018national}, which requires user registration and adherence to specific data use policies. For this study, the authors utilized data from SHHS Visit 1, which includes two EEG channels, two EOG channels, and one EMG channel, among other signals. The recordings are manually scored according to the R\&K guidelines. For comparability with literature approaches \cite{phan2023lseqsleepnet}, recordings that did not include all stages—W, NREM stages N1 through N4, and REM—were excluded. This resulted in the retention of 5463 recordings. As with SEDF, the study merged N3 and N4, and excluded data segments labeled as MOVEMENT and UNKNOWN. The study conducted experiments on multi-channel data containing two EEG channels (C4-A1 and C3-A2), two EOG channels (left and right), and one EMG channel, as well as on single-channel data consisting only of the C4-A1 EEG channel. the authors performed a random split of the recordings, allocating 70\% of the recordings for training and 30\% for testing, following the split ratio used in \cite{phan2023lseqsleepnet}. Additionally, 100 recordings from the training set were set aside specifically for validation purposes. Again, these splits are released for later reproducibility as part of the accompanying code repository \cite{repo}.

\header{Rationale for dataset selection} The study selected three datasets of varying sizes to align with its research focus effectively. The medium-sized SEDF allowed the authors to conduct the entire training-testing process for dozens of different models throughout the model selection phase. This comprehensive evaluation, encompassing a broad range of models and hyperparameter settings, would have been infeasible on the much larger SHHS1 dataset. Secondly, and equally important, utilizing three independent datasets allowed the study to rigorously test the generalizability of the identified models across diverse data sources. This approach mitigates the risk of overfitting to a specific dataset, thereby ensuring that the achieved findings are robust and applicable to various clinical settings and populations.

\header{Preprocessing} The channels utilized across all datasets were standardized to a uniform sample rate of 100 Hz using band-limited interpolation \cite{Smith2015DigitalAudioResampling}, unless the original signal frequency was already at 100 Hz. All datasets were used in terms of two different input representations, raw time series and spectrograms. Spectrograms were generated in a manner consistent with L-SeqSleepNet \cite{phan2023lseqsleepnet}. Each channel signal from a 30-second epoch was transformed to generate a log-magnitude time-frequency image. This involved dividing the signal into 29 time steps and 129 frequency bins using the short-time Fourier transform (STFT) with a window length of 2 seconds and a 1 second overlap. A Hamming window and a 256-point fast Fourier transform (FFT) were used in the process. The resulting amplitude spectrum for each channel underwent logarithmic transformation, creating a time-frequency image. The raw time series input underwent no further preprocessing. A systematic investigation of the impact of additional preprocessing steps, such as filtering, is beyond the scope of this work.

\section{Training procedure and performance evaluation}
\label{app:train_and_eval}
\header{Training procedure} 
To mitigate issues due to imbalanced label distributions that are typical for this task, focal loss \cite{lin2017focal} was used as the loss function.
For all experiments, the authors used a fixed effective batch size of 64 achieved through gradient accumulation. The study performed experiments with learning rates determined via the learning rate finder \cite{smith2017cyclical} as well as with a fixed learning rate of 0.001 for each experiment. The learning rate yielding superior validation set performance was selected. The authors trained models using a constant learning rate and AdamW \cite{loshchilov2018decoupled} as the optimizer. For training and validation, they divided the whole input sequence of a given sample into consecutive segments whose lengths coincide with the model's input size. For SEDF and MASS-SS3, the models underwent training for 50 epochs, with validation performed after each training epoch. For SHHS1, longer training was found to be beneficial (in terms of training iterations) and correspondingly models were trained for 30 epochs. For the final test set evaluation, the model at the training epoch with the best validation set performance (in terms of macro-$F_1$ score, introduced in detail below) was selected. The study trained on non-overlapping crops of the specified input size. During test time, the study split the input sequences into segments of the same length but use a smaller stride length coinciding with the length of a single epoch. For most segments, the study obtained in this way multiple predictions corresponding to different positions in the input sequence passed to the model. Subsequently, all available predictions at the level of output probabilities for a given segment were averaged.

\header{Performance metrics} Irrespective of the input size, the procedure described in the previous section yields a probabilistic model prediction per test/validation set epoch. One can then compare dichotomized model predictions with the ground truth. The most commonly used metric in the field is the macro-$F_1$ score, computed as the mean of the individual label $F_1$ scores,
\begin{equation}
\text{macro}\,F_1 = \frac{1}{5} \sum_{c\in C} F_{1,c}\,,
\end{equation}
where $C=\{W,N1,N2,N3,REM\}$ and $F_{1,c}$ is the $F_1$ score for class $c$. The macro $F_1$ score serves as the primary target metric since model selection on the validation score is carried out based on this metric. For informativeness, the authors report a range of other metrics but stress that the models have not been specifically optimized for these metrics. In particular, they report the individual label $F_1$ scores that constitute the macro $F_1$ score and the overall prediction accuracy but rush to add that the use of accuracy as a target metric is problematic in the presence of label imbalance and is therefore most sensitive to the performance on the majority class of NREM class N2. Finally, the study also reports the macro average of the area under the receiver operating curve (AUROC) defined as
\begin{equation}
\text{macro} \text{AUROC}= \frac{1}{5} \sum_{c\in C} {\text{AUROC}}_{c}\,,
\end{equation}
where $C=\{W,N1,N2,N3,REM\}$ and ${\text{AUROC}}_{c}$ denotes the area under the receiver operating curve for class $c$, considering class $c$ as positive and the remaining classes as negative classes. This metric is an example for a commonly used scoring metric that is based on (the ranking of) output probabilities rather than dichotomized outputs and therefore providing complementary information compared to metrics in the latter category.

\header{Score estimation}
The authors based the study's most reliable experiments on the SEDF(full) dataset and the SHHS1 dataset employing a straightforward holdout set evaluation scheme. The exception to this approach is the small-scale SEDF20 subset and MASS-SS3 dataset, where the study used a  cross-validation scheme to conform to the predominantly used evaluation scheme in the literature. For full transparency, the authors indicate the precise dataset choice as well as the score estimation method for all considered methods in the final results in Table~\ref{tab:mainresult_sedf} and Table~\ref{tab:mainresult_massshhs}.

\header{Predictive uncertainty during model selection}
An uncertainty estimate for the achieved results is very desirable already during model selection to assess the significance of the performance differences. There are two main sources of uncertainty that are of  interest: the uncertainty in the final score due to the randomness of the training process and the uncertainty due to the finite size and specific composition of the test/validation set. The former uncertainty is typically assessed through multiple training runs. For computational reasons, the study resorted to a single training run at this stage and deferred a proper assessment to the evaluation in Section~\ref{sec:testseteval}. However, the study assessed the second kind of uncertainty through (empirical) bootstrapping ($n=1000$ iterations) on the test set. To determine if the difference in performance (according to the macro-F1 score as metric) is significant, the study calculated 95\% confidence intervals for the score difference. If the confidence interval for the performance difference does not overlap with zero, the study regarded the performances of the two models as statistically significantly different.

\header{Predictive uncertainty during final model evaluation}
To assess the systematic uncertainty due to the randomness of the training process, the study performed three training runs per scenario and reported the median $F_1$-score as well as the interquartile ranges as a measure of fluctuations. The authors estimated the statistical uncertainty due to the finiteness and composition of the test set through empirical bootstrapping ($n=1000$ iterations) on the test set. The study reports the median of the three bootstrap uncertainty measures across the three training runs. The final results are reported in the form $(\text{median macro}-F_1)\pm(\text{systematic uncertainty})\pm(\text{statistical uncertainty})$. To assess the statistical significance of performance differences while also incorporating the uncertainty due to the randomness of the training process, the study follows the methodology proposed in \cite{Mehari2022}. Three training runs for model A and model B yield a total of 9 combinations. Bootstrapping the score difference as described above allows assessment of the statistical significance of the score difference between pairs of models individually. The study then set a threshold related to the willingness to accept fluctuations due to the randomness of the training process, and declare model A superior to model B if the fraction of pairwise tests with significant outcomes exceeds the predetermined threshold. As in \cite{Mehari2022}, this study used a threshold of 60\%.

\section{Additional results}

\subsection{Epoch encoder vs. sub-epoch encoder}
In \Cref{tab:subee}, the study investigates the effect of using epoch encoders that operate on fractions of an entire input epoch.

\begin{table}[!!!h]
	\centering
	\captionsetup{justification=centering, singlelinecheck=false, position=top}
	\caption{Comparing epoch encoder and sub-epoch encoder on SEDF based on validation set scores.}
	\label{tab:subee}
	\scriptsize
	\begin{tabular}{p{1.8cm}p{1.8cm}p{1.5cm}p{2.0cm}p{2.0cm}p{2.3cm}}
		\toprule
		    channels&features&fraction&encoder&predictor&macro-$F_1$\\ 
\midrule 
		\multirow{8}{*}{single}&\multirow{4}{*}{raw}
		&\textbf{1}&EES4&S4&\textbf{0.801}\\        
		&&\textbf{1/2}&EES4&S4&\textbf{0.802}\\
		&&\textbf{1/5}&EES4&S4&\textbf{0.804}\\
		&&1/10&EES4&S4&0.784\\
		
		\cline{2-6}
							
		&\multirow{4}{*}{spec}
		&\textbf{1}&EENS4&S4&\textbf{0.800}\\      
		&&{1/2}&EENS4&S4&0.787\\
		&&1/5&EENS4&S4&0.795\\
		&&1/10&EENS4&S4&0.792\\
                 		
		\midrule 

		\multirow{8}{*}{multi}&\multirow{4}{*}{raw}		
		&\textbf{1}&EES4&S4&\textbf{0.820}\\   
		&&1/2&EES4&S4&0.814\\
		&&\textbf{1/5}&EES4&S4&\textbf{0.820}\\
		&&1/10&EES4&S4&0.814\\
							
		\cline{2-6}
							
		&\multirow{4}{*}{spec}
		&\textbf{1}&EENS4&S4&\textbf{0.815}\\
		&&1/2&EENS4&S4&0.812\\
		&&1/5&EENS4&S4&0.810\\
		&&\textbf{1/10}&EENS4&S4&\textbf{0.816}\\

		\bottomrule	
	\end{tabular}
\parbox{\linewidth}{\tiny Bold text highlights the sub-epoch fraction that achieves the highest scores, considering the associated uncertainties.}
\end{table}

\subsection{Single-epoch models on SEDF}
Table~\ref{tab:mainresult_singleepoch} shows the performance of the best-performing single-epoch models from Table~\ref{tab:sedf1} trained and evaluated on the SEDF dataset (test set scores). These results provide quantitative insights on the performance gains achieved through predicting several epochs at once.

\begin{table}[!!!h]
	\centering

	\captionsetup{justification=centering, singlelinecheck=false, position=top}
	\caption{Performance of selected models using a single epoch as input on SEDF. Notation as in Table~\ref{tab:15epoch}. }
	\label{tab:mainresult_singleepoch}
	
	\begin{tabular}{lll}
	
		\toprule
			model & channels  & macro-$F_1$\\
		
		\midrule 
		
		 \textbf{CNN+S4(TS)} & single &\textbf{0.752$\pm$0.003$\pm$0.002} \\
		 None+S4(Spec) & single &  0.730$\pm$0.007$\pm$0.004 \\
		
        \cline{1-3}
       
		 CNN+S4(TS) & multi & 0.769$\pm$0.002$\pm$0.004 \\
		 \textbf{None+S4(Spec)} & multi & \textbf{0.780$\pm$0.002$\pm$0.003}  \\

		\bottomrule	
	\end{tabular}

\end{table}

\end{document}